\documentclass[journal,twoside,web]{ieeecolor}
\usepackage{generic}
\usepackage{cite}
\usepackage{amsmath,amssymb,amsfonts}
\usepackage{algorithmic}
\usepackage{graphicx}
\usepackage{hyperref}
\hypersetup{hidelinks=true}
\usepackage{textcomp}
\usepackage{multirow}
\usepackage{color}
\usepackage{amssymb}
\usepackage{makecell}
\usepackage{comment}
\usepackage{subfigure}
\usepackage{bm}

\def\BibTeX{{\rm B\kern-.05em{\sc i\kern-.025em b}\kern-.08em
    T\kern-.1667em\lower.7ex\hbox{E}\kern-.125emX}}
\begin{document}
\title{Completed Feature Disentanglement Learning for Multimodal MRIs Analysis}

\author{Tianling Liu, Hongying Liu, Fanhua Shang, Lequan Yu, Tong Han, Liang Wan
	\thanks{Received 2 April 2024; revised 6 January 2025; accepted 31 January 2025. This work was supported in part by the National Natural Science Foundation of China under Grant 62476196, Grant 62276182 and Grant 82071994, in part by the Hong Kong Innovation and Technology Fund under Grant ITS/273/22, in part by Peng Cheng Lab Program under Grant PCL2023A08, in part by Tianjin Municipal Education Commission Research Plan under Grant 2024ZX008, in part by Health Industry High-level Talent Selection and Training Project Foundation under Grant TJSJMYXYC-D2-059, in part by Tianjin Health Science and Technology Project Foundation under Grant TJWJ2024RC016. \emph{(Corresponding author: Lequan Yu and Liang Wan.)}}
	\thanks{Tianling Liu, Fanhua Shang are with the College of Intelligence and Computing, Tianjin University, Tianjin, China (e-mail: liu\_dling@tju.edu.cn, fhshang@tju.edu.cn).}
        \thanks{Liang Wan is with the College of Intelligence and Computing, Tianjin University, and also with the Medical School of Tianjin University, Tianjin, China (e-mail: lwan@tju.edu.cn).}
	\thanks{Hongying Liu is with the Medical School of Tianjin University, Tianjin, and also with the Peng Cheng Lab, Shenzhen, China (email: hyliu2009@tju.edu.cn).}
	\thanks{Lequan Yu is with the Department of Statistics and Actuarial Science, School of Computing and Data Science, The University of Hong Kong, Hong Kong SAR, China (e-mail: lqyu@hku.hk).}
	\thanks{Tong Han is with the Department of Radiology, Tianjin Huanhu Hospital, and also with the Tianjin Key Laboratory of Cerebral Vascular and Neurodegenerative Diseases, Tianjin, China (e-mail: mrbold@163.com).}
	\thanks{The code is available at https://github.com/IsDling/CFDL.}
}

\maketitle

\begin{abstract}
Multimodal MRIs play a crucial role in clinical diagnosis and treatment. Feature disentanglement (FD)-based methods, aiming at learning superior feature representations for multimodal data analysis, have achieved significant success in multimodal learning (MML). Typically, existing FD-based methods separate multimodal data into modality-shared and modality-specific features, and employ concatenation or attention mechanisms to integrate these features. 
However, our preliminary experiments indicate that these methods could lead to a loss of shared information among subsets of modalities when the inputs contain more than two modalities, and such information is critical for prediction accuracy. Furthermore, these methods do not adequately interpret the relationships between the decoupled features at the fusion stage.
To address these limitations, we propose a novel Complete Feature Disentanglement (CFD) strategy that recovers the lost information during feature decoupling. Specifically, the CFD strategy not only identifies modality-shared and modality-specific features, but also decouples shared features among subsets of multimodal inputs, termed as modality-partial-shared features. We further introduce a new Dynamic Mixture-of-Experts Fusion (DMF) module that dynamically integrates these decoupled features, by explicitly learning the local-global relationships among the features. The effectiveness of our approach is validated through classification tasks on three multimodal MRI datasets. Extensive experimental results demonstrate that our approach outperforms other state-of-the-art MML methods with obvious margins, showcasing its superior performance.
\end{abstract}


\section{Introduction}
\label{sec:introduction}
\begin{figure}[!t]
    \centering
    \includegraphics[width=0.85\linewidth]{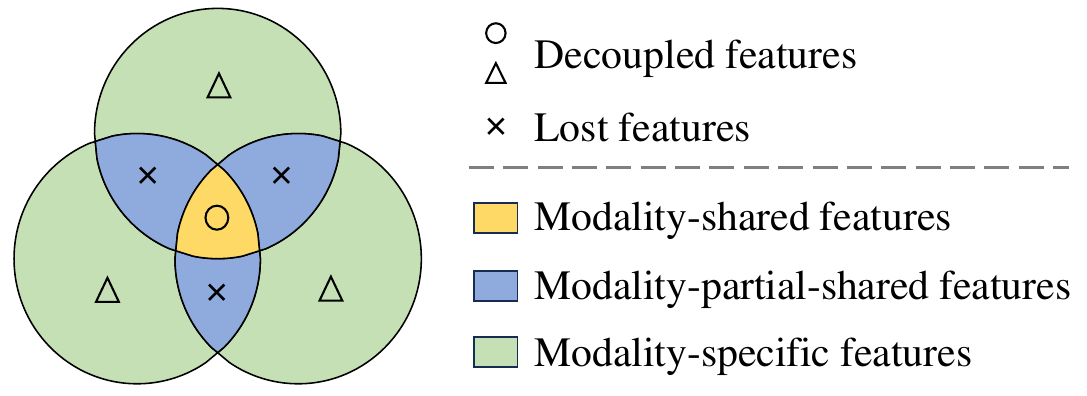}
    \caption{
    A concept map for illustrations of incomplete feature representations in existing FD methods in the three-modal case.}

    \label{completed_information_threemodals_concept}
\end{figure}

\begin{figure}[!t]
    \centering
    \includegraphics[width=0.9\linewidth]{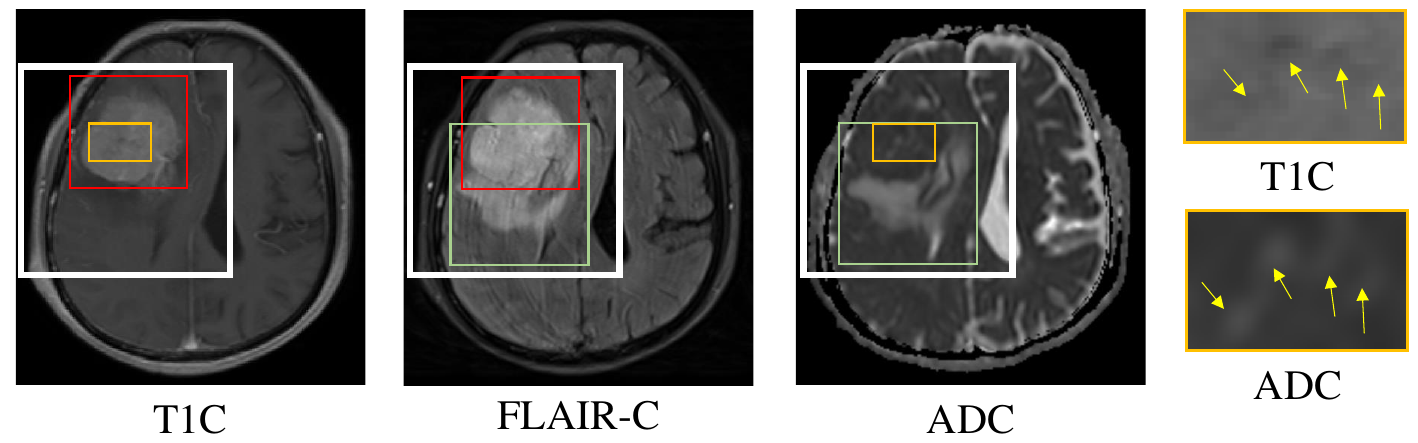}
    \caption{
    An example from the MEN dataset for illustrations of information loss between pair-wise modalities. Region of interest (ROI), including the tumor, edema and their surrounding area, is marked with an white box and used as the input of the model. Tumor and edema areas are marked with red and green boxes, respectively, and the yellow box highlights regions of high cell density in the tumor.
    Tumor characteristics are shared between T1C and FLAIR-C, edema characteristics between FLAIR-C and ADC, and cell density information between T1C and ADC.
    }

    \label{completed_information_threemodals_case}
\end{figure}


Multi-modality data contains multiple aspects of information about an object, and different modalities can provide complementary information. Numerous previous studies have demonstrated the remarkable success of multimodal learning (MML)~\cite{baltruvsaitis2018multimodal}
for medical image analysis. 
However, inappropriate processing of multimodal information can significantly impact the efficiency of MML. 
According to~\cite{huang2021makes}, the key to successful MML lies in achieving a higher quality of feature representation. 
Many previous works~\cite{he2021multi,braman2021deep,zhou2019deep,ning2021relation,hu2020disentangled,hazarika2020misa,cheng2021multimodal,li2023decoupled,zheng2022multi} have focused on enhancing the learning performance, which can be classified into three categories. 
Two categories focus on extracting the shared information between multiple modalities~\cite{braman2021deep,he2021multi} or specific information of each modality~\cite{zhou2019deep,ning2021relation}. These methods cannot fully extract multimodal information, as they only focus on one type of feature, leading to information loss~\cite{zheng2022multi,li2023decoupled}. The third category focuses on feature disentanglement (FD), which
decouples modality-shared features as well as modality-specific features~\cite{hu2020disentangled,cheng2021multimodal,hazarika2020misa,li2023decoupled}, leading to sound results. 

We revisit the relationship between the representation spaces of multimodal data. As illustrated in the concept map of the three-modal case in Fig.~\ref{completed_information_threemodals_concept}, intuitively, we can consider that there exists modality-shared information (yellow area) as well as modality-specific information (green area). Upon further consideration and exploration, we discover that there is shared information present between subsets of modalities (blue area). However, existing FD methods could potentially ignore such information. 
On the other hand, our preliminary experiments reveal that the lost information is crucial for accurate prediction (see the first and fourth rows of Table~\ref{as_men}). 
In Fig.~\ref{completed_information_threemodals_case}, we take 
MEN dataset as an example for illustration. 
Both T1C and FLAIR-C highlight the tumor area, indicating shared tumor information. Similarly, FLAIR-C and ADC highlight the edema area, showing shared edema information. Additionally, T1C and ADC share information about cell density in the tumor area.
In fact, such shared information among pair-wise modalities is found to be relevant to the prediction of meningiomas grade and invasion in clinical research~\cite{hess2018brain,li2019presurgical,chen2023radiotherapy}.
%
%
%

Furthermore, current multimodal fusion studies mainly focus on uncertainty-based fusion methods~\cite{han2020trusted,han2022trusted,liu2024dynamic} and attention-based methods~\cite{zhang2021modality,xing2022nestedformer}. 
However, these approaches generally address modality-shared and modality-specific information fusion across multiple modalities, while overlooking the design of fusion mechanisms for shared information between subsets of modalities.

To tackle the above issues, we propose a completed feature disentanglement multimodal learning (CFDL) approach for multimodal MRIs analysis.
First, we present a novel completed feature disentanglement (CFD) strategy to address the information loss in previous FD-based methods. In addition to decoupling modality-shared features among all modalities and modality-specific features, We further decouple features shared between subsets of modalities, referred to as \textit{modality-partial-shared features}.
The modality partial-shared features are also expected to have higher similarities while being dissimilar from the other two kinds of features.
As demonstrated in Section \uppercase\expandafter{\romannumeral 4}, these features play a critical role in prediction performance.
%
Next, to improve the interpretability of feature fusion, we propose a new dynamic mixture-of-experts fusion (DMF) module, which can explicitly capture local-global interrelationships between the decoupled features to achieve more effective fusion.
Finally, we evaluate our framework on three multimodal MRI datasets, demonstrating its effectiveness and superiority compared to state-of-the-art methods.

\begin{figure*}[t]
    \centering
    \includegraphics[width=0.9\linewidth]{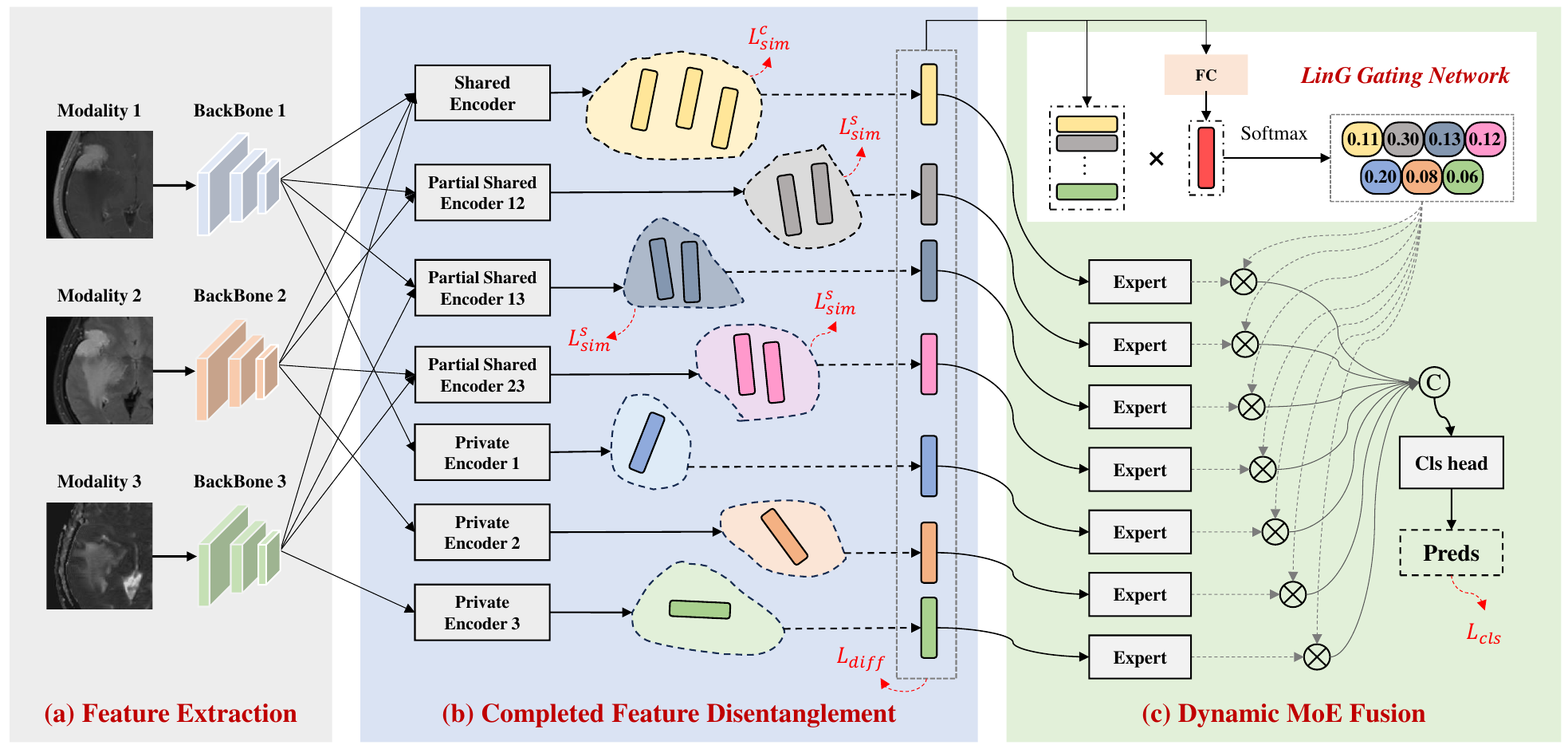}
    \caption{The overview of proposed CFDL framework in the three-modal case. (a) For each modality, we adopt the same type of backbone for feature extraction. 
    (b) Completed Feature Disentanglement (CFD) strategy decouples each extracted features to modality-shared features, modality-specific features, and modality-partial-shared features between pair-wise modalities. (c) Dynamic MoE Fusion (DMF) module dynamically and appropriately fuses decoupled features based on samples benefitted from LinG\_GN. LinG\_GN can obtain the complex interrelationships between these features. Specifically, \textcircled{c} means concatenation operation, $FC$ represents a fully-connected layer, $ClS\ head$ denotes a classification head.}
    \label{framework}
\end{figure*}
\section{Related Work}


\subsection{Feature Representation Learning in MML}
Feature representation learning is a crucial aspect in the field of MML, which contains three types of methods. Some approaches~\cite{braman2021deep,he2021multi} have focused on the first type of method, which extracted specific features from each modality and subsequently fused them with obtained embeddings. Braman~\emph{et al.}~\cite{braman2021deep} designed the Multimodal Orthogonalization (MMO) loss function to obtain the maximum specific representation for each of radiology, pathology, genomic and clinical data. 
Several methods~\cite{zhou2019deep,ning2021relation} have concentrated on the second type of method, which captured modality-shared features from multiple modalities. Ning~\emph{et al.}~\cite{ning2021relation} built a bi-directional mapping between original space and shared space of multimodal to effectively obtained multi-modal shared representation.
However, the fist two type of methods have primarily emphasized either modality-specific or modality-shared features, thus failing to learn a comprehensive representation of multimodal data. 
The third type of method, FD, has proven to be effective in separating multimodal information into meaningful components and has been successfully applied in various applications~\cite{hu2020disentangled,cheng2021multimodal,hazarika2020misa,li2023decoupled}.
Hu~\emph{et al.}~\cite{hu2020disentangled} proposed a disentangled-multimodal adversarial autoencoder (DMM-AAE) model that employed a VAE to disentangle 
multimodal MRIs information into modal-common features and modal-specific features.
However, this method only addressed the two-modal fusion scenario. Cheng~\emph{et al.}~\cite{cheng2021multimodal} extended this approach  to multimodal fusion scenario. It is worth noting that both of these methods cannot be trained end-to-end due to the reliance on hand-crafted features as model inputs. Hazarika~\emph{et al.}~\cite{hazarika2020misa} decoupled multimodal information into modality-invariant and modality-specific features using Central Moment Discrepancy metric, orthogonality constraints and reconstruction loss. Li~\emph{et al.}~\cite{li2023decoupled} proposed decoupled multimodal distillation (DMD), which first separated the representation of each modality into modality-irrelevant space and modality-exclusive space. Then, a graph distillation unit was employed to each space for dynamic enhancing the features of each modality.


The aforementioned FD methods have a common drawback that can result in incomplete feature representation learning in the case of three or more modalities, as depicted in Fig.~\ref{completed_information_threemodals_concept}. 
In contrast, the proposed CFD strategy addresses this limitation by decoupling multimodal information into modality-shared features, modality-specific features, and modality-partial-shared features, thereby enabling comprehensive feature representation learning.

\subsection{Multimodal Feature Fusion}
The fusion strategy is another crucial aspect of MML. Several approaches have involved concatenating features extracted from different modalities~\cite{gao2020mgnn,he2021multi}
or representation spaces~\cite{hu2020disentangled,cheng2021multimodal,li2023decoupled}, such as modality-shared and modality-specific features.
However, concatenation fusion has not effectively utilized the correlations between multiple modalities.

In recent years, there has been an increasing focus on exploring the correlations among multiple modalities to obtain effective features.
Some methods achieve multimodal fusion by assigning weights or probabilities to each modality. 
\cite{han2020trusted,han2022trusted,liu2024dynamic} explored the uncertainties of different modalities to obtain reliable multimodal fusion information.
Choi~\emph{et al.}~\cite{choi2019embracenet} proposed EmbraceNet which performed  multimodal representation fusion based on a probabilistic approach. 
Zhou~\emph{et al.}~\cite{zhou2023attentive} introduced a canonical correlation analysis (CCA)-based method named ADCCA to exploit the correlation between multiple modalities and integrate the complementary information from these modalities.
Zhuang~\emph{et al.}~\cite{zhuang2024glomo} proposed a global-guided fusion method which consider both global and local correlations of multiple modalities.

With the proven ability of attention mechanisms to enhance feature representation and explore complex correlation between multiple modalities, many attention-based multimodal fusion methods have emerged~\cite{li2022adaptive,zhang2021modality,zhu2022multimodal,xing2022nestedformer,zhou2023coco}. Zhang~\emph{et al.}~\cite{zhang2021modality} proposed a modality-aware mutual learning (MAML) framework that weighted the multimodal features using an attention-based modality-aware (MA) module. 
Zhu~\emph{et al.}~\cite{zhu2022multimodal} captured complementary information from multimodal data using self-attention and cross-modal attention, and further designed a triple network to obtain more discriminative information. Xing~\emph{et al.}~\cite{xing2022nestedformer} developed the NestedFormer framework, 
which included the Nested Modality-aware Feature Aggregation (NMaFA) module to explore long-range correlations within and between modalities for effective and comprehensive information learning. However, these attention-based methods cannot explicitly reveal the contribution of each decoupled features during fusion process.

Mixture-of-Experts (MoE)~\cite{Jacobs1991Adaptive} employs multiple experts to extract distinct representation spaces from the input and generated corresponding weights using a gating network. MoE have the ability to dynamically capture the mixture information from multiple experts. Several studies~\cite{goyal2016multimodal,cao2023multi} have extended MoE to handle multi-input scenarios, where each expert processes a specific input. These approaches leverage the dynamic nature of MoE. However, these methods concatenate all inputs to generate weights in the gating network without thoroughly considering the relationships between different inputs, which can limit the effectiveness of the fusion process. In contrast, we introduce a gating network to capture the local-global relationships between the decoupled features.
\section{Method}
Let's denote the input multimodal data as $\{X^i,y^i\}_{i=1}^N$, where $N$ is the number of samples; $X^i=\{x^i_j\}_{j=1}^M$, $M$ denotes the modality number of each sample; $y^i$ is the classification label for the $i$-th sample.

\subsection{Overview}
The proposed CFDL framework, illustrated in Fig.~\ref{framework}, comprises three parts: a) feature extraction from multimodal MRIs, b) completed feature disentanglement for feature decoupling, and c) dynamic MoE fusion for dynamically integrating the decoupled features. The framework employs the same type of backbone to extract latent features $x_j$ from each modality. 
To capture a comprehensive representation of the multimodal data, the latent features are decoupled into modality-shared features, modality-specific features, and modality-partial-shared features using the proposed CFD strategy. The decoupled features are then integrated via the DMF module.
Within the DMF module, each decoupled feature is paired with a specific expert, and a gating network named LinG\_GN generates weights for multiple experts. The fused features are obtained by aggregating the weighted features from the experts. 
In the following, we take three-modal condition as examples to illustrate the CFD strategy and DMF module.

\subsection{Completed Feature Disentanglement Strategy}
\label{sec:CFD}


Inspired by previous FD methods~\cite{hu2020disentangled,hazarika2020misa,cheng2021multimodal}, we first decouple the extracted latent features into modality-shared features and modality-specific features. We employ a shared-encoder $\bm{E^{c}}$ to decouple modality-shared features and three private-encoders $\bm{E^{p}}_j$ to decouple modality-specific feature for each modality.
Three modality-shared features can be formulated as follows:
\begin{equation}
F_j = \bm{E^{c}}(x_j),
\end{equation}
and three modality-specific features can be obtained with:
\begin{equation}
P_j = \bm{E^{p}}_j(x_j).
\end{equation}
The \emph{final modality-shared feature} $F$ is the mean of all modality-shared features, given by
\begin{equation}
F\ = \frac{1}{M} \sum_{j} F_j.
\end{equation}

We further consider modality-partial-shared features between pair-wise modalities. As a result, three groups of modality-partial-shared features are decoupled, with each group consisting of two features. The two features in the same group ($G_{jk}^j$, $G_{jk}^k$) are decoupled with the same partial-shared-encoder named $\bm{E^{s}}_{jk}$, i.e.,
\begin{equation}
G_{jk}^j = \bm{E^{s}}_{jk}(x_j), \quad
G_{jk}^k = \bm{E^{s}}_{jk}(x_k).
\end{equation}
Specifically, $G_{jk}^j$ represents the modality-partial-shared feature between the $j$-th modality and the $k$-th modality, which is decoupled from the $j$-th modality. The \emph{final modality-partial-shared feature} $G_{jk}$ can be calculated by averaging the modality-partial-shared features in each group,
\begin{equation}
G_{jk} = \frac{1}{2} (G_{jk}^j+G_{jk}^k).
\end{equation}

In the end, we obtain three modality-shared features ($F_1$, $F_2$, $F_3$), three modality-specific features ($P_1$, $P_2$, $P_3$), and three groups of modality-partial-shared features including $\{G_{12}^1, G_{12}^2\}$, $\{G_{13}^1, G_{13}^3\}$, $\{G_{23}^2, G_{23}^3\}$. Furthermore, we get 7 \emph{final decoupled features}, which is denoted as a set $\mathcal{S}$, where $\mathcal{S} = \{F, P_1, P_2, P_3, G_{12}, G_{13}, G_{23}\}$.

To enhance the completeness of the decoupled representation, we use the following three constraints:
\begin{itemize}
	\item[1)] Modality-shared features should exhibit high similarity to one another.
	\item[2)] Modality-partial-shared features within each group should have maximum similarity.
	\item[3)] The \emph{final decoupled features} should exhibit maximum dissimilarity from one another.
\end{itemize}

To ensure the effective decoupling of modality-shared features and modality-partial-shared features, we employ the mean squared error (MSE) loss as a constraint. 
The MSE loss measures the discrepancy between two features, and we aim to increase similarity between two features by minimizing this loss. The losses for modality-shared features $ \mathcal{L}_{sim}^c$ and modality-partial-shared features $\mathcal{L}_{sim}^s$ are expressed as:
\begin{equation}
\mathcal{L}_{sim}^c = \sum_{j=1} \sum_{k=j+1} MSE(F_j,F_k),
\end{equation}
\begin{equation}
\mathcal{L}_{sim}^s = \sum_{j=1} \sum_{k=j+1} MSE(G_{jk}^j,G_{jk}^k).
\end{equation}
We denote $\mathcal{L}_{sim}$ as the sum of $\mathcal{L}_{sim}^c$ and $\mathcal{L}_{sim}^s$.

To enhance the decoupling of modality-specific features and minimize redundancy among all \emph{final decoupled features}, we incorporate cosine similarity as a constraint for better optimization. 
Our objective is to increase the dissimilarity between the final decoupled features by reducing the cosine similarity between each pair of these features.
The loss for all final decoupled features $L_{diff}$ is calculated by:
\begin{equation}
\mathcal{L}_{diff} = \sum_{j=1} \sum_{k=j+1} CS(\mathcal{S}_j, \mathcal{S}_k),
\end{equation}
where $CS(\cdot,\cdot)$ represents the cosine similarity function.

\subsection{Dynamic MoE Fusion Module}

To ensure dynamic fusion of the final decoupled features, we introduce the DMF module based on the MoE architecture, which is shown in Fig.~\ref{framework} (c).
In DMF module, each final decoupled feature $\mathcal{S}_j \in \mathbb{R}^{dim}$ is associated with a specific expert $\bm{E^x}_j$, which is implemented as a fully-connected layer.
We introduce LinG\_GN to dynamically generate weights for these experts, taking into account the relationships of these final decoupled features. The LinG\_GN operates with two inputs to capture a comprehensive understanding of these features.
Firstly, we concatenate all the final decoupled features as one input for the LinG\_GN. To integrate this concatenated feature, we map the concatenated feature into a unified representation utilizing a fully-connected layer $FC$. This unified representation captures the collective information from all the final decoupled features and is treated as the global feature $g\in\mathbb{R}^{dim}$,
\begin{equation}
g = FC(ReLU(Cat(\mathcal{S}_1,...,\mathcal{S}_7))),
\end{equation}
where $Cat(\cdot,...,\cdot)$ means column concatenation operation and $ReLU$ is the activation function.
Secondly, we stack all the final decoupled features as the second input for the LinG\_GN. This stacked features retain the individual information from each final decoupled feature and are considered as the local features, denoted as $\textbf{O}=[{\mathcal{S}_1}^T,...,{\mathcal{S}_7}^T]$, where $\textbf{O} \in\mathbb{R}^{7\times dim}$. By obtaining both the global feature and the local features, we can explore the importance of each local feature within the context of the global feature. This exploration allows us to determine the weight $\omega$ of local features in contributing to the fused representation. The weight $\omega$ is calculated by:
\begin{equation}
\omega = softmax(\textbf{O} \times g),
\end{equation}
where $\times$ represents matrix multiplication operation, $\omega\in \mathbb{R}^{7}$. Each element in $\omega$ represents the weight for the corresponding expert network $\bm{E^x}_j$. 
The fused feature $F_{fused}$ is obtained by concatenating the linearly weighted experts,
\begin{equation}
F_{fused} = Cat(\omega_1 \bm{E^x}_1(\mathcal{S}_1),...,\omega_7 \bm{E^x}_{7}(\mathcal{S}_7)).
\end{equation}

For the final prediction $\hat{y}$, we utilize a Multi-Layer Perceptron ($MLP$) as the classifier. The cross-entropy (CE) loss is employed as the supervision for prediction. The classification loss $\mathcal{L}_{cls}$ is defined as:
\begin{equation}
\mathcal{L}_{cls} = CE(\hat{y},y).
\end{equation}

The final loss $\mathcal{L}$ can be defined as the weighted sum of aforementioned losses,
\begin{equation}
\mathcal{L} = \mathcal{L}_{cls} + \alpha\mathcal{L}_{sim}+ \beta \mathcal{L}_{diff},
\end{equation}
where $\alpha$ and $\beta$ are balance factors.

\subsection{Network Architecture}
We utilize the 3D ResNet18~\cite{he2016deep} as the backbone for feature extraction, and the parameters of these backbones are not shared. The dimension of the each extracted latent feature $x_j$ is 512. The shared-encoder $\bm{E^c}$, each private-encoders $\bm{E^p}_j$, each partial-shared-encoders $\bm{E^s}_{jk}$, $FC$
and each expert $\bm{E^x}_j$ are all implemented as one fully-connected layer with $dim$ neurons. The $MLP$ consists of two fully-connected layers with $dim$ neurons and one output layer with $num\_cls$ neurons, where $num\_cls$ represents the number of classes. Each fully-connected layer in $MLP$ is followed by a ReLU layer and a Dropout layer. We empirically set $dim$ as 32.

\begin{figure}[t]
	\centering
	\includegraphics[width=\linewidth]{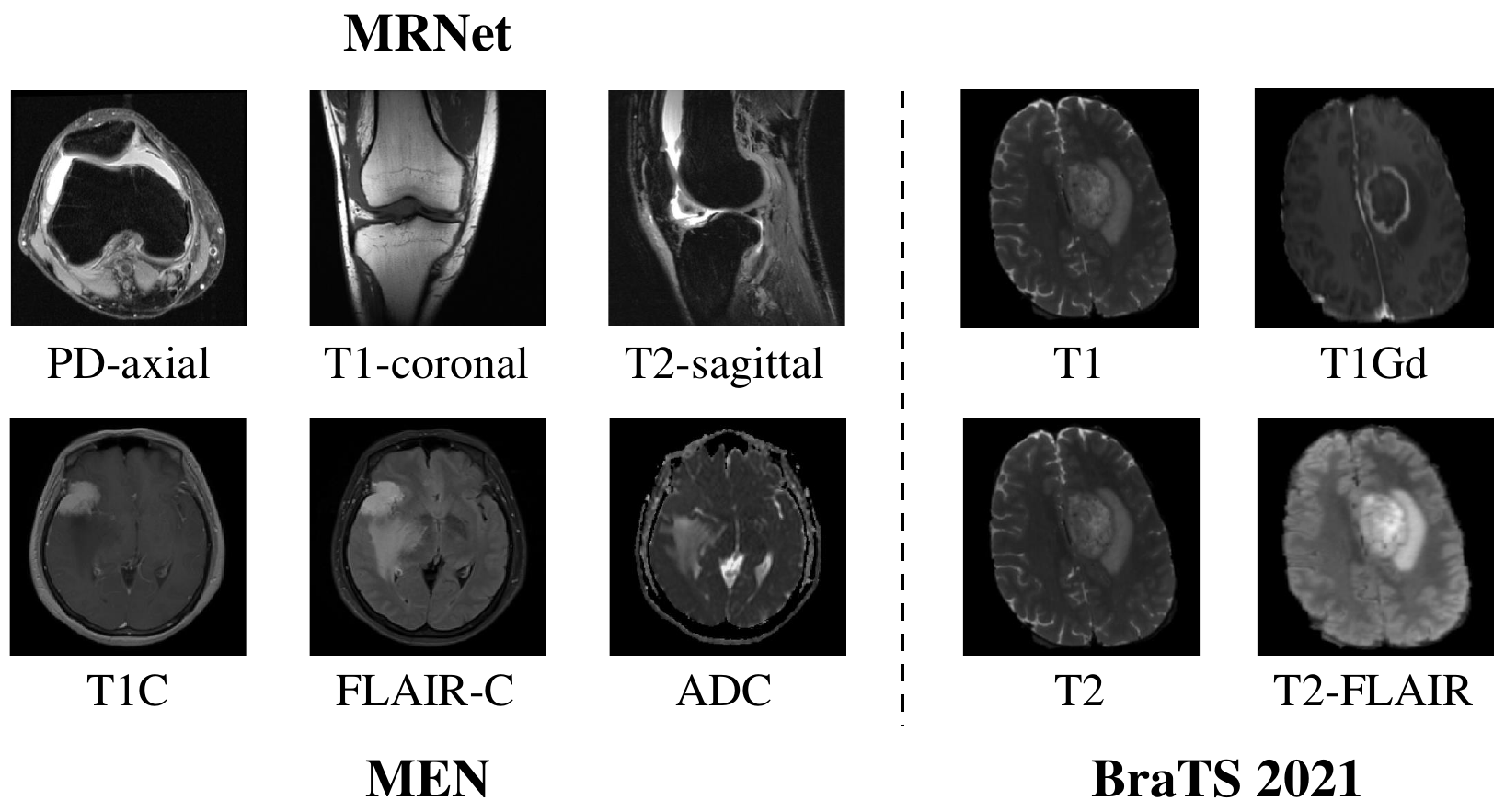}
	\caption{The display of cases from three multimodal MRI datasets.}
	\label{data_show_onecase}
\end{figure}

\section{Experiments}
\subsection{Datasets and Tasks}
We utilize three multimodal MRI datasets, which consist of two public datasets and one private dataset, to verify the effectiveness of the proposed framework. Some cases from these three datasets are shown in Fig~\ref{data_show_onecase}. 

\subsubsection{Meniscal Tear Prediction}
For the prediction of meniscal tear, we employ the MRNet dataset~\cite{bien2018deep}, which is a publicly available knee multi-MRI dataset\footnote{More information about the MRNet dataset can be available in https://stanfordmlgroup.github.io/competitions/mrnet/.}. 
There are 1130 cases in training set and 120 cases in validation set. The training set contains 397 meniscal tear cases and 733 contrast cases, and the validation set contains 52 meniscal tear cases and 68 contrast cases. Each case includes three MRIs: sagittal plane T2-weighted series (T2-sagittal), coronal plane T1-weighted series (T1-coronal) and axial plane PD-weighted series (PD-axial). 
We resize the MRIs to 24*128*128 as the model input.

\subsubsection{Meningiomas Grading Prediction}
We collect Meningiomas Grading Prediction Dataset, referred to as MEN, from the Brain Medical Center of Tianjin University, Tianjin Huanhu Hospital \footnote{The Ethical Committee of Tianjin Huanhu Hospital approves scientific research using these MRIs and waives the need for informed patient consent ((Jinhuan) Ethical Review No.(2022-046)).}.
This dataset consists of three grades of meningiomas: Grade 1 (G1), Grade 2 with invasion (G2inv) and Grade 2 without invasion (G2ninv). The total dataset comprises 798 cases, including 650 Grade\_1, 62 Grade\_2inv and 86 Grade\_2ninv cases. Each case includes three brain MRIs: Contrast-Enhanced T1 series (T1C), Contrast-Enhanced T2 FLAIR series (FLAIR-C) and Apparent Diffusion Coefficient series (ADC). Following previous works~\cite{adeli2018prediction,joo2021extensive}, we request radiologists to crop the regions of interest (ROIs). To maintain the shape of the tumor and edema regions, the ROIs are zero-padded into squares and resized to dimensions of 24*128*128, which serve as the inputs for the model.

\subsubsection{MGMT Promoter Status Prediction}
The MGMT Promoter Status Prediction Dataset, known as BraTS 2021~\cite{baid2021rsna}, is a publicly available multimodal brain MRI dataset. It encompasses cases with MGMT methylated (MGMT+) and unmethylated (MGMT-) status. The dataset comprises 580 available cases\footnote{
We drop 5 cases from original 585 cases during pre-processing. There are 3 cases with unexpected issues. The other 2 cases cannot be registered by CaPTk. Get more information in https://www.kaggle.com/c/rsna-miccai-brain-tumor-radiogenomic-classification.}, with each case containing four modalities: T1, post-contrast T1-weighted (T1Gd), T2-weighted (T2), and T2 Fluid Attenuated Inversion Recovery (T2-FLAIR). Specifically, there are 275 MGMT- cases and 305 MGMT+ cases. Pre-processing for each modality involves image registration and skull-stripping using the Cancer Imaging Phenomics Toolkit (CaPTk)~\cite{davatzikos2018cancer}. We crop the ROIs using masks generated by the pretrained Swin UNETR~\cite{hatamizadeh2021swin}. Finally, we zero-pad the ROIs into squares and resize them to 16*128*128 as the inputs for the proposed method.

\subsection{Implementation Details}

\subsubsection{Training Details}
We employ 3-fold cross-validation for private MEN and public BraTS 2021 datasets, and train three times using different seeds with already divided training and validation data for MRNet dataset. During model training, 
we implement several techniques to prevent overfitting, such as data augmentation, L2 regularization (weight decay) and dropout~\cite{srivastava2014dropout}. Data augmentation techniques include random clip, random crop, gaussian noise and random erasing~\cite{zhong2020random}. The weight decay is set as $1e-4$, and the dropout value is set to $0.5$. The network is optimized with the Adam optimizer~\cite{kingma2014adam}. We linearly warm up the learning rate from zero to the preset value over $5$ epochs and apply a learning rate decay strategy, reducing the learning rate to $0.8$ after every $5$ epochs. The batch size is set $32$. For MRNet dataset, we initialize the learning rate value as $8e-4$, and set the number of epochs to $50$. 
For MEN dataset, the learning rate is specified as $5e-4$, and number of epochs is fixed as $100$. 
For BraTS 2021 dataset, we preset the learning rate value as $2e-4$ and the number of epochs as $50$.
The balance factors, $\alpha$ and $\beta$, are set to \{$0.5, 0.005$\}, \{$1, 1$\}, \{$0.001, 0.001$\} for MRNet, MEN and BraTS, respectively. 
Details of the ablation analysis for the balance factors are provided in the supplementary material.
All experiments are conducted with PyTorch on an NVIDIA RTX 3090 GPU.

\subsubsection{Evaluation Metrics}
For two-class datasets, MRNet and BraTS 2021, we employ seven metrics to assess the effectiveness of the proposed framework, including Sensitivity (SEN), Specificity (SPE), Accuracy (ACC), G-mean, Balanced Accuracy (Ba\_ACC)~\cite{brodersen2010balanced}, Area Under the Precision-Recall Curve (AUPRC), Area Under the Curve (AUC). For three-class dataset, MEN, we utilize seven evaluation metrics, including Accuracy (ACC), Accuracy of G1 (ACC\_G1), Accuracy of G2inv (ACC\_G2inv), Accuracy of G2ninv (ACC\_G2ninv), weighted F1 score (weighted-F1), macro F1 score (macro-F1) and AUC. For the statistical analysis, Wilcoxon signed-rank~\cite{wilcoxon1992individual} is adopted to compare the metrics of our proposed framework with other methods.

\subsubsection{Compared Methods}
We compare the proposed framework with nine state-of-the-art (SOTA) MML methods, including EmbraceNet~\cite{choi2019embracenet}, ETMC~\cite{han2022trusted}, ADCCA~\cite{zhou2023attentive}, MAML~\cite{zhang2021modality}, NestedFormer~\cite{xing2022nestedformer}, MISA~\cite{hazarika2020misa}, DMD~\cite{li2023decoupled}, CCML~\cite{liu2024dynamic} and GLoMo~\cite{zhuang2024glomo}. Specifically, ETMC and CCML are uncertainty-based MML methods. 
ADCCA and GLoMo are correlation-based MML methods.
MAML and NestedFormer are attention-based MML methods originally designed for segmentation, but adapted to the classification task by adding a classifier after the encoders. MISA and DMD are FD-based MML methods. 
To ensure a fair comparison, in addition to the transformer-based NestedFormer method, we set the backbones of other CNN-based comparison methods to be the same as that of the proposed framework.


\begin{figure}[t]
\centering
    \subfigure[The proposed framework with $\alpha=0$, $\beta=0$ on the MRNet dataset]{%
        \includegraphics[width=0.47\linewidth]{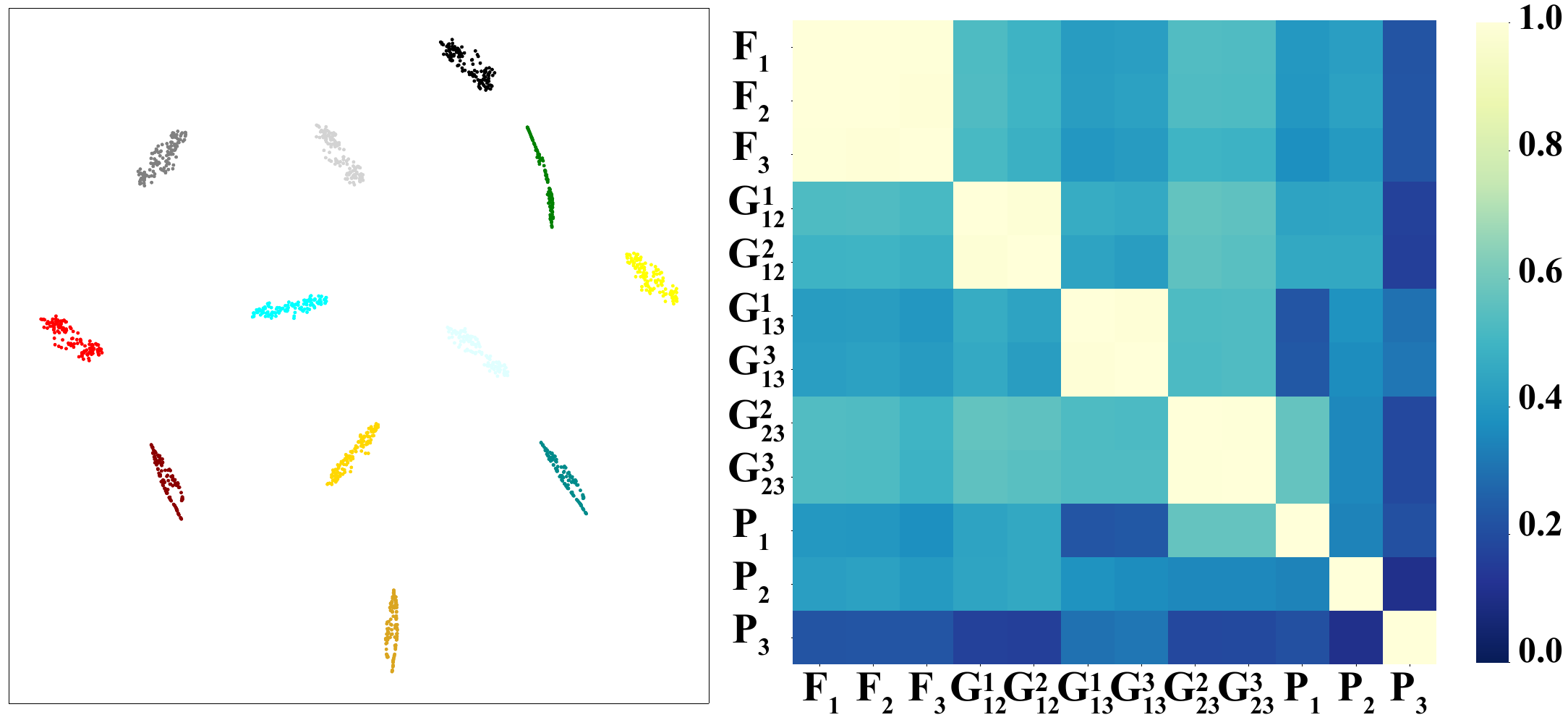}}
    \quad
    \subfigure[The proposed framework with $\alpha\neq0$, $\beta\neq0$ on the MRNet dataset]{%
        \includegraphics[width=0.47\linewidth]{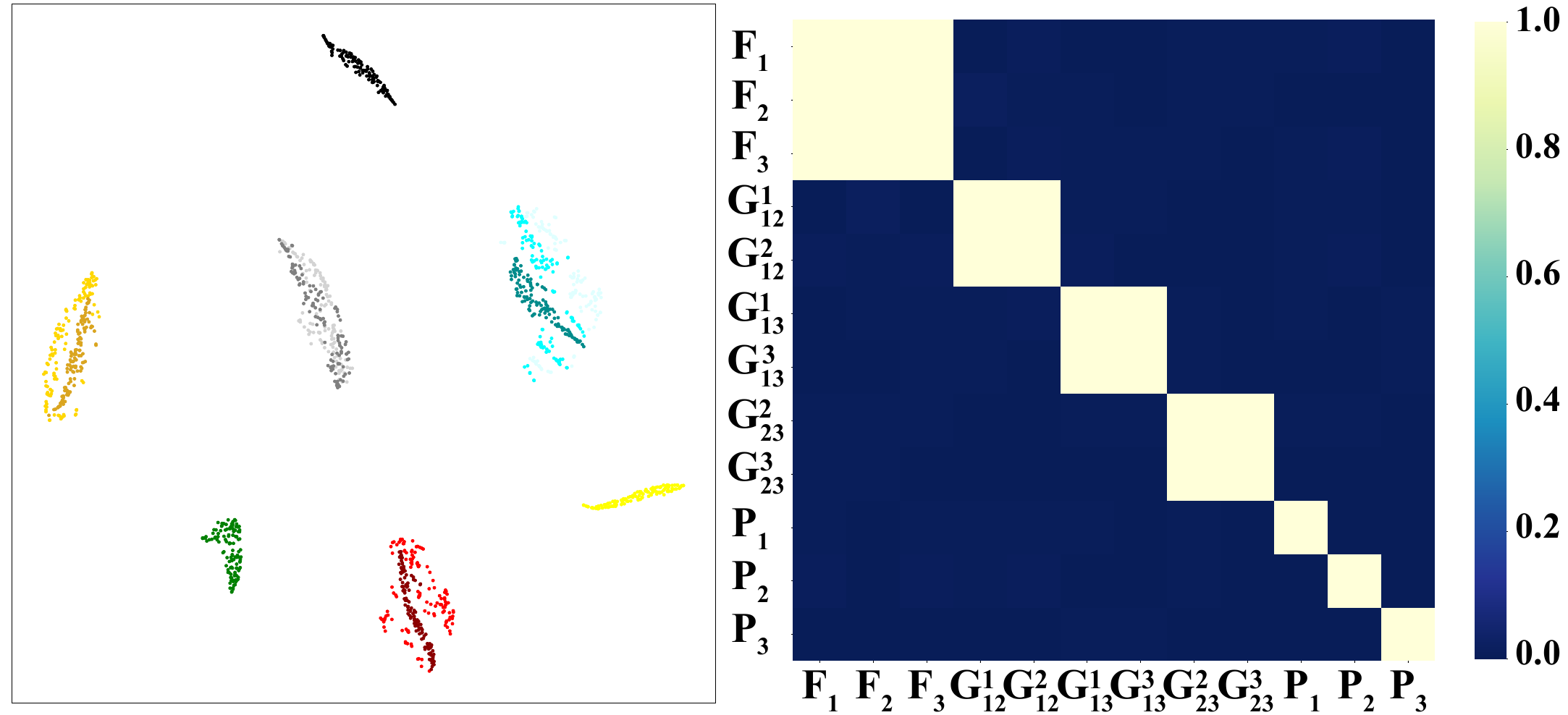}}
    \subfigure[The proposed framework with $\alpha=0$, $\beta=0$ on the MEN dataset]{%
        \includegraphics[width=0.47\linewidth]{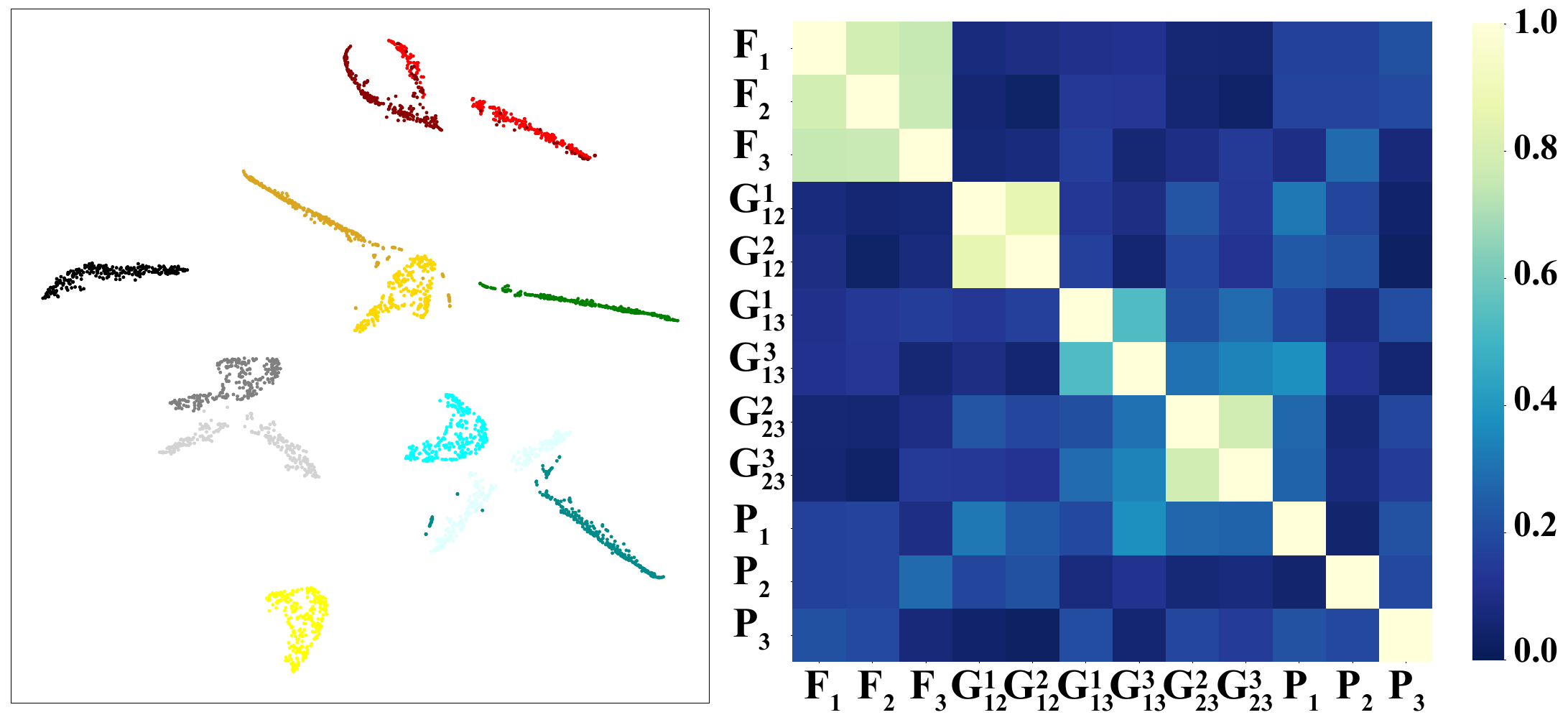}}
    \quad
    \subfigure[The proposed framework with $\alpha\neq0$, $\beta\neq0$ on the MEN dataset]{%
        \includegraphics[width=0.47\linewidth]{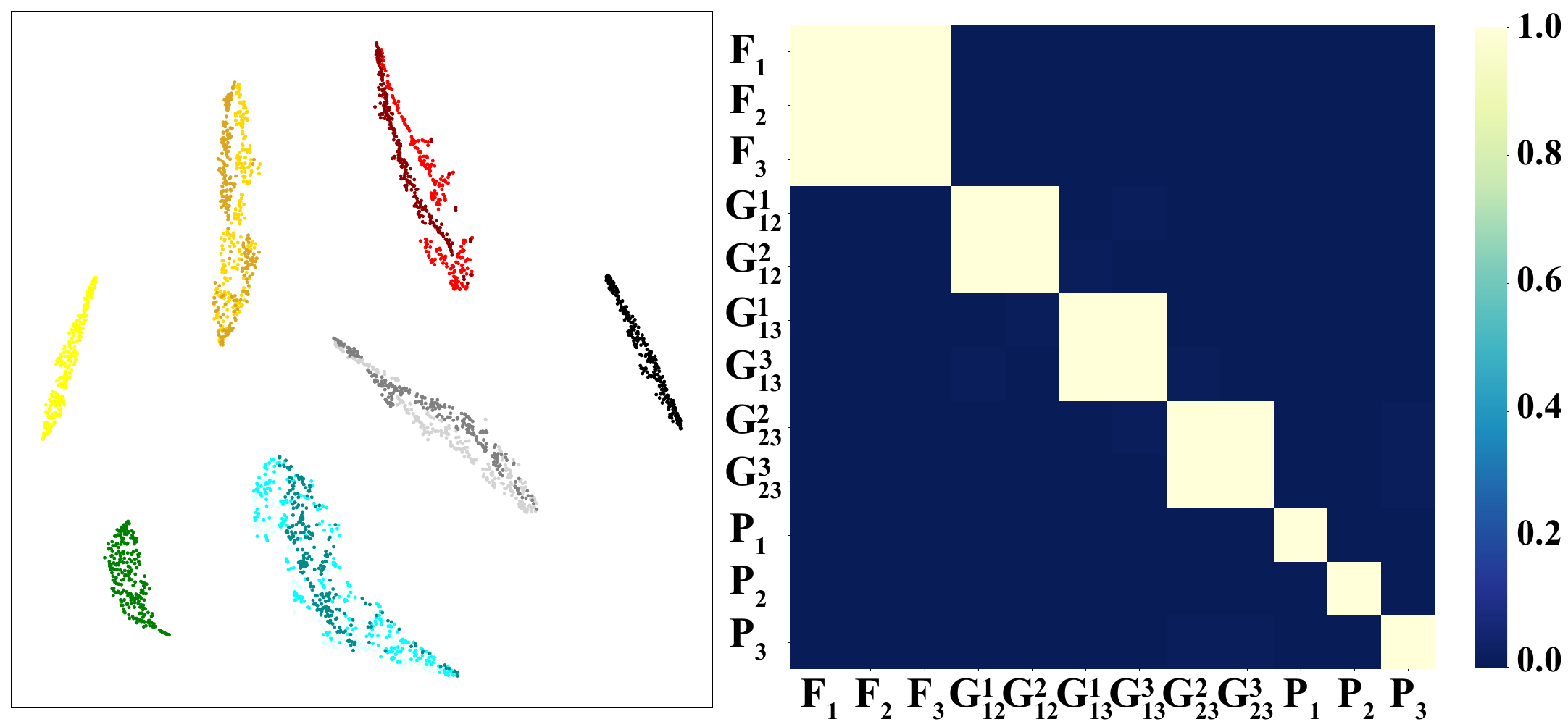}}
    \subfigure{%
        \includegraphics[width=0.75\linewidth]{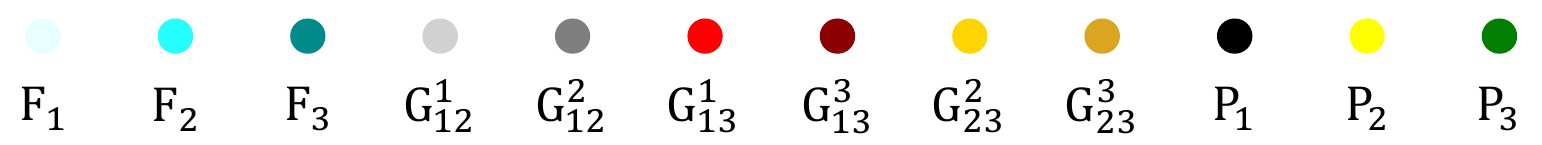}}
%
\caption{
Visualization results of the proposed framework on the MRNet and MEN datasets. Each sub-figure includes t-SNE visualization of the decoupled features (left) and a heatmap of the cosine similarities for each pair-wise decoupled features (right). In the heatmap, yellow indicates higher similarity, while blue indicates lower similarity.
}
\label{visual}
\end{figure}

\begin{table}[t]
\centering
\caption{The comparison results on the MRNet dataset (mean$\pm$standard deviation). The best and second best results for each metric are respectively highlighted by \textcolor{red}{red} and \textcolor{blue}{blue}. The marker ``*" indicates that our proposed framework achieves statistically significant results than other methods (p-value<0.05).}
\label{comparison_mrnet}
\Large
\resizebox{0.5\textwidth}{!}{
\begin{tabular}{cccccccc}
\hline
Method       & SEN                                                      & SPE                                                      & ACC                                                      & G-Mean                                                  & Ba\_ACC                                                  & AUPRC                                                    & AUC                                                      \\ \hline
EmbraceNet~\cite{choi2019embracenet}   & \makecell[c]{\vspace{-1mm} 0.6410*\\ \large $ \pm $ 0.0111} & \makecell[c]{\vspace{-1mm} \textcolor{blue}{0.7696}\\ \large $ \pm $ 0.0424} & \makecell[c]{\vspace{-1mm} 0.7139*\\ \large $ \pm $ 0.0192} & \makecell[c]{\vspace{-1mm} 0.7021*\\ \large $ \pm $ 0.0136} & \makecell[c]{\vspace{-1mm} 0.7053*\\ \large $ \pm $ 0.0157} & \makecell[c]{\vspace{-1mm} 0.5924*\\ \large $ \pm $ 0.0199} & \makecell[c]{\vspace{-1mm} 0.7509*\\ \large $ \pm $ 0.0136} \\
MISA~\cite{hazarika2020misa}         & \makecell[c]{\vspace{-1mm} 0.6795*\\ \large $ \pm $ 0.0111} & \makecell[c]{\vspace{-1mm} 0.7451\\ \large $ \pm $ 0.0557} & \makecell[c]{\vspace{-1mm} 0.7167*\\ \large $ \pm $ 0.0289} & \makecell[c]{\vspace{-1mm} 0.7111*\\ \large $ \pm $ 0.0232} & \makecell[c]{\vspace{-1mm} 0.7123*\\ \large $ \pm $ 0.0249} & \makecell[c]{\vspace{-1mm} 0.5963*\\ \large $ \pm $ 0.0314} & \makecell[c]{\vspace{-1mm} 0.7517*\\ \large $ \pm $ 0.0294} \\
MAML~\cite{zhang2021modality}         & \makecell[c]{\vspace{-1mm} 0.6923\\ \large $ \pm $ 0.1071} & \makecell[c]{\vspace{-1mm} 0.7598\\ \large $ \pm $ 0.1390} & \makecell[c]{\vspace{-1mm} \textcolor{blue}{0.7306}\\ \large $ \pm $ 0.0337} & \makecell[c]{\vspace{-1mm} 0.7183*\\ \large $ \pm $ 0.0200} & \makecell[c]{\vspace{-1mm} \textcolor{blue}{0.7261*}\\ \large $ \pm $ 0.0186} & \makecell[c]{\vspace{-1mm} \textcolor{blue}{0.6148}\\ \large $ \pm $ 0.0346} & \makecell[c]{\vspace{-1mm} 0.7609*\\ \large $ \pm $ 0.0068} \\
ETMC~\cite{han2022trusted}         & \makecell[c]{\vspace{-1mm} \textcolor{red}{0.7949}\\ \large $ \pm $ 0.0675} & \makecell[c]{\vspace{-1mm} 0.6274*\\ \large $ \pm $ 0.0946} & \makecell[c]{\vspace{-1mm} 0.7000*\\ \large $ \pm $ 0.0300} & \makecell[c]{\vspace{-1mm} 0.7031*\\ \large $ \pm $ 0.0268} & \makecell[c]{\vspace{-1mm} 0.7111*\\ \large $ \pm $ 0.0224} & \makecell[c]{\vspace{-1mm} 0.5834*\\ \large $ \pm $ 0.0266} & \makecell[c]{\vspace{-1mm} 0.7655*\\ \large $ \pm $0.0169} \\
NestedFormer~\cite{xing2022nestedformer} & \makecell[c]{\vspace{-1mm} 0.7308\\ \large $ \pm $ 0.0193} & \makecell[c]{\vspace{-1mm} 0.7206*\\ \large $ \pm $ 0.0255} & \makecell[c]{\vspace{-1mm} 0.7250*\\ \large $ \pm $ 0.0167} & \makecell[c]{\vspace{-1mm} \textcolor{blue}{0.7255*}\\ \large $ \pm $ 0.0159} & \makecell[c]{\vspace{-1mm} 0.7257*\\ \large $ \pm $ 0.0160} & \makecell[c]{\vspace{-1mm} 0.6042*\\ \large $ \pm $ 0.0174} & \makecell[c]{\vspace{-1mm} 0.7540*\\ \large $ \pm $ 0.0113} \\
ADCCA~\cite{zhou2023attentive} & \makecell[c]{\vspace{-1mm} \textcolor{blue}{0.7500}\\ \large $ \pm $ 0.0693} & \makecell[c]{\vspace{-1mm} 0.6765*\\ \large $ \pm $ 0.0589} & \makecell[c]{\vspace{-1mm} 0.7083*\\ \large $ \pm $ 0.0084} & \makecell[c]{\vspace{-1mm} 0.7104*\\ \large $ \pm $ 0.0096} & \makecell[c]{\vspace{-1mm} 0.7133*\\ \large $ \pm $ 0.0094} & \makecell[c]{\vspace{-1mm} 0.5882*\\ \large $ \pm $ 0.0073} & \makecell[c]{\vspace{-1mm} 0.7627*\\ \large $ \pm $ 0.0050} \\
DMD~\cite{li2023decoupled}          & \makecell[c]{\vspace{-1mm} 0.6154*\\ \large $ \pm $ 0.1201} & \makecell[c]{\vspace{-1mm} \textcolor{red}{0.7990}\\ \large $ \pm $ 0.0945} & \makecell[c]{\vspace{-1mm} 0.7195*\\ \large $ \pm $ 0.0048} & \makecell[c]{\vspace{-1mm} 0.6955*\\ \large $ \pm $ 0.0244} & \makecell[c]{\vspace{-1mm} 0.7072*\\ \large $ \pm $ 0.0136} & \makecell[c]{\vspace{-1mm} 0.5997*\\ \large $ \pm $ 0.0051} & \makecell[c]{\vspace{-1mm} 0.7668*\\ \large $ \pm $ 0.0112} \\
CCML~\cite{liu2024dynamic}          
& \makecell[c]{\vspace{-1mm} 0.6795*\\ \large $ \pm $ 0.0728} 
& \makecell[c]{\vspace{-1mm} \textcolor{blue}{0.7696}\\ \large $ \pm $ 0.0473} 
& \makecell[c]{\vspace{-1mm} \textcolor{blue}{0.7306}\\ \large $ \pm $ 0.0096} 
& \makecell[c]{\vspace{-1mm} 0.7214*\\ \large $ \pm $ 0.0199} 
& \makecell[c]{\vspace{-1mm} 0.7245*\\ \large $ \pm $ 0.0153} 
& \makecell[c]{\vspace{-1mm} 0.6100*\\ \large $ \pm $ 0.0100} 
& \makecell[c]{\vspace{-1mm} 0.7669*\\ \large $ \pm $ 0.0085} \\
GLoMo~\cite{zhuang2024glomo}         
& \makecell[c]{\vspace{-1mm} 0.6667*\\ \large $ \pm $ 0.0867} 
& \makecell[c]{\vspace{-1mm} 0.7451\\ \large $ \pm $ 0.1251} 
& \makecell[c]{\vspace{-1mm} 0.7111*\\ \large $ \pm $ 0.0337} 
& \makecell[c]{\vspace{-1mm} 0.6995*\\ \large $ \pm $ 0.0165} 
& \makecell[c]{\vspace{-1mm} 0.7059*\\ \large $ \pm $ 0.0199} 
& \makecell[c]{\vspace{-1mm} 0.5934*\\ \large $ \pm $ 0.0344} 
& \makecell[c]{\vspace{-1mm} \textcolor{blue}{0.7670*}\\ \large $ \pm $ 0.0154} \\ \hline
Proposed     & \makecell[c]{\vspace{-1mm} 0.7244\\ \large $ \pm $ 0.0588} & \makecell[c]{\vspace{-1mm} 0.7500\\ \large $ \pm $ 0.0778} & \makecell[c]{\vspace{-1mm} \textcolor{red}{0.7389}\\ \large $ \pm $ 0.0255} & \makecell[c]{\vspace{-1mm} \textcolor{red}{0.7351}\\ \large $ \pm $ 0.0178} & \makecell[c]{\vspace{-1mm} \textcolor{red}{0.7372}\\ \large $ \pm $ 0.0199} & \makecell[c]{\vspace{-1mm} \textcolor{red}{0.6207}\\ \large $ \pm $ 0.0301} & \makecell[c]{\vspace{-1mm} \textcolor{red}{0.8029}\\ \large $ \pm $ 0.0219} \\ \hline
\end{tabular}}
\end{table}
\begin{table}[t]
\centering
\caption{The comparison results on the MEN dataset.
}
\label{comparison_men}
\Large
\resizebox{0.5\textwidth}{!}{
\begin{tabular}{cccccccc}
\hline
Method       & ACC                                                & \makecell[c]{\vspace{-1mm} ACC\\ \_G1}                                            & \makecell[c]{\vspace{-1mm} ACC\\ \_G2inv}                                         & \makecell[c]{\vspace{-1mm} ACC\\ \_G2ninv}                                      & \makecell[c]{\vspace{-1mm} weighted\\ -F1}                                       & \makecell[c]{\vspace{-1mm} macro\\ -F1}                                          & AUC                                                \\ \hline
EmbraceNet~\cite{choi2019embracenet}   & \makecell[c]{\vspace{-1mm} 0.9136*\\ \large $ \pm $  0.0103} & \makecell[c]{\vspace{-1mm} 0.9570\\ \large $ \pm $  0.0206} & \makecell[c]{\vspace{-1mm} 0.7894*\\ \large $ \pm $  0.0353} & \makecell[c]{\vspace{-1mm} 0.6730*\\ \large $ \pm $  0.1192} & \makecell[c]{\vspace{-1mm} 0.9155*\\ \large $ \pm $  0.0100} & \makecell[c]{\vspace{-1mm} 0.8053*\\ \large $ \pm $  0.0329} & \makecell[c]{\vspace{-1mm} 0.8683*\\ \large $ \pm $  0.0286} \\
MISA~\cite{hazarika2020misa}         & \makecell[c]{\vspace{-1mm} 0.9099*\\ \large $ \pm $  0.0103} & \makecell[c]{\vspace{-1mm} \textcolor{blue}{0.9678}\\ \large $ \pm $  0.0199} & \makecell[c]{\vspace{-1mm} 0.8848\\ \large $ \pm $  0.0603} & \makecell[c]{\vspace{-1mm} 0.4873*\\ \large $ \pm $  0.0535} & \makecell[c]{\vspace{-1mm} 0.9056*\\ \large $ \pm $  0.0074} & \makecell[c]{\vspace{-1mm} 0.7885*\\ \large $ \pm $  0.0301} & \makecell[c]{\vspace{-1mm} 0.9411*\\ \large $ \pm $  0.0092} \\
MAML~\cite{zhang2021modality}         & \makecell[c]{\vspace{-1mm} \textcolor{blue}{0.9360*}\\ \large $ \pm $  0.0107} & \makecell[c]{\vspace{-1mm} \textcolor{red}{0.9707}\\ \large $ \pm $  0.0209} & \makecell[c]{\vspace{-1mm} \textcolor{blue}{0.9015}\\ \large $ \pm $  0.0523} & \makecell[c]{\vspace{-1mm} 0.6992*\\ \large $ \pm $  0.1059} & \makecell[c]{\vspace{-1mm} \textcolor{blue}{0.9361*}\\ \large $ \pm $  0.0076} & \makecell[c]{\vspace{-1mm} \textcolor{blue}{0.8524*}\\ \large $ \pm $  0.0229} & \makecell[c]{\vspace{-1mm} 0.9600*\\ \large $ \pm $  0.0113} \\
ETMC~\cite{han2022trusted}         & \makecell[c]{\vspace{-1mm} 0.8834*\\ \large $ \pm $  0.0088} & \makecell[c]{\vspace{-1mm} 0.9385*\\ \large $ \pm $  0.0138} & \makecell[c]{\vspace{-1mm} 0.8045*\\ \large $ \pm $  0.0569} & \makecell[c]{\vspace{-1mm} 0.5198*\\ \large $ \pm $  0.1783} & \makecell[c]{\vspace{-1mm} 0.8836*\\ \large $ \pm $  0.0158} & \makecell[c]{\vspace{-1mm} 0.7595*\\ \large $ \pm $  0.0512} & \makecell[c]{\vspace{-1mm} 0.9051*\\ \large $ \pm $  0.0364} \\
NestedFormer~\cite{xing2022nestedformer} & \makecell[c]{\vspace{-1mm} 0.9273*\\ \large $ \pm $  0.0060} & \makecell[c]{\vspace{-1mm} 0.9584\\ \large $ \pm $  0.0167} & \makecell[c]{\vspace{-1mm} 0.8212*\\ \large $ \pm $  0.0620} & \makecell[c]{\vspace{-1mm} \textcolor{blue}{0.7682*}\\ \large $ \pm $  0.0468} & \makecell[c]{\vspace{-1mm} 0.9298*\\ \large $ \pm $  0.0036} & \makecell[c]{\vspace{-1mm} 0.8436*\\ \large $ \pm $  0.0030} & \makecell[c]{\vspace{-1mm} \textcolor{blue}{0.9695*}\\ \large $ \pm $  0.0083} \\
ADCCA~\cite{zhou2023attentive} & \makecell[c]{\vspace{-1mm} 0.8295*\\ \large $ \pm $  0.0446} & \makecell[c]{\vspace{-1mm} 0.9091*\\ \large $ \pm $  0.0698} & \makecell[c]{\vspace{-1mm} 0.5803*\\ \large $ \pm $  0.1253} & \makecell[c]{\vspace{-1mm} 0.4119*\\ \large $ \pm $  0.2221} & \makecell[c]{\vspace{-1mm} 0.8284*\\ \large $ \pm $  0.0301} & \makecell[c]{\vspace{-1mm} 0.6289*\\ \large $ \pm $  0.0534} & \makecell[c]{\vspace{-1mm} 0.8753*\\ \large $ \pm $  0.0421} \\
DMD~\cite{li2023decoupled}          & \makecell[c]{\vspace{-1mm} 0.8897*\\ \large $ \pm $  0.0076} & \makecell[c]{\vspace{-1mm} 0.9677\\ \large $ \pm $  0.0002} & \makecell[c]{\vspace{-1mm} 0.8030*\\ \large $ \pm $  0.1046} & \makecell[c]{\vspace{-1mm} 0.3619*\\ \large $ \pm $  0.0644} & \makecell[c]{\vspace{-1mm} 0.8798*\\ \large $ \pm $  0.0101} & \makecell[c]{\vspace{-1mm} 0.7303*\\ \large $ \pm $  0.0279} & \makecell[c]{\vspace{-1mm} 0.9202*\\ \large $ \pm $  0.0227} \\
CCML~\cite{liu2024dynamic}          
& \makecell[c]{\vspace{-1mm} 0.8550*\\ \large $ \pm $  0.0419} 
& \makecell[c]{\vspace{-1mm} 0.9187*\\ \large $ \pm $  0.0607} 
& \makecell[c]{\vspace{-1mm} 0.8106*\\ \large $ \pm $  0.1344} 
& \makecell[c]{\vspace{-1mm} 0.4024*\\ \large $ \pm $  0.2119} 
& \makecell[c]{\vspace{-1mm} 0.8551*\\ \large $ \pm $  0.0270} 
& \makecell[c]{\vspace{-1mm} 0.7003*\\ \large $ \pm $  0.0255} 
& \makecell[c]{\vspace{-1mm} 0.9164*\\ \large $ \pm $  0.0257} \\
GLoMo~\cite{zhuang2024glomo}          
& \makecell[c]{\vspace{-1mm} 0.9059*\\ \large $ \pm $  0.0240} 
& \makecell[c]{\vspace{-1mm} 0.9507\\ \large $ \pm $  0.0416} 
& \makecell[c]{\vspace{-1mm} 0.8697*\\ \large $ \pm $  0.0605} 
& \makecell[c]{\vspace{-1mm} 0.5952*\\ \large $ \pm $  0.0899} 
& \makecell[c]{\vspace{-1mm} 0.9066*\\ \large $ \pm $  0.0184} 
& \makecell[c]{\vspace{-1mm} 0.7968*\\ \large $ \pm $  0.0211} 
& \makecell[c]{\vspace{-1mm} 0.9585*\\ \large $ \pm $  0.0213} \\
\hline
Proposed     & \makecell[c]{\vspace{-1mm} \textcolor{red}{0.9462}\\ \large $ \pm $  0.0113} & \makecell[c]{\vspace{-1mm} 0.9616\\ \large $ \pm $  0.0160} & \makecell[c]{\vspace{-1mm} \textcolor{red}{0.9182}\\ \large $ \pm $  0.0315} & \makecell[c]{\vspace{-1mm} \textcolor{red}{0.8492}\\ \large $ \pm $  0.0383} & \makecell[c]{\vspace{-1mm} \textcolor{red}{0.9483}\\ \large $ \pm $  0.0101} & \makecell[c]{\vspace{-1mm} \textcolor{red}{0.8936}\\ \large $ \pm $  0.0106} & \makecell[c]{\vspace{-1mm} \textcolor{red}{0.9776}\\ \large $ \pm $  0.0021} \\ \hline
\end{tabular}}
\end{table}

\subsection{Quantitative Results}
\subsubsection{Evaluation on Meniscal Tear Prediction Dataset}
The comparison results on the MRNet dataset are summarized in Table~\ref{comparison_mrnet}. Among the comparison methods, the correlation-based method GLoMo~\cite{zhuang2024glomo} achieves the best AUC ($0.7670$), while the uncertainty-based method CCML~\cite{liu2024dynamic} obtains the second place ($0.7669$). The attention-based method MAML~\cite{zhang2021modality} obtains better results than other comparison methods in three metrics: ACC ($0.7306$), Ba\_ACC ($0.7261$) and AUPRC ($0.6148$). Our proposed framework achieves the first place in five metrics: ACC ($0.7389$, $0.0083$ better than the 2nd), G-Mean ($0.7351$, $0.0096$ better than the 2nd), Ba\_ACC ($0.7372$, $0.0111$ better than the 2nd), AUPRC ($0.6207$, $0.0059$ better than the 2nd) and AUC ($0.8029$, $0.0359$ better than the 2nd). The higher ACC and AUC, along with the more balanced accuracy between positive and negative cases, demonstrate the effectiveness of proposed framework on the MRNet dataset. 

\subsubsection{Evaluation on Meningiomas Grading Prediction Dataset}
We further validate the proposed framework on the private MEN dataset, and the comparison results are shown in Table~\ref{comparison_men}. Among the comparison methods, the attention-based method NestedFormer achieves the best AUC ($0.9695$) and ACC\_G2ninv ($0.7682$). The other attention-based method, MAML, achieves better rusults than other comparison methods in five metrics, including ACC ($0.9360$), ACC\_G1 ($0.9707$), ACC\_G2inv ($0.9015$), weighted-F1 ($0.9361$) and macro-F1 ($0.8524$). 
The reason for the poor performance of DMD is that the difficult prediction of class G2ninv results in distillation leaning towards other classes.  
In contrast, our proposed framework achieves first place in six metrics: ACC ($0.9462$, $0.0102$ better than the 2nd), ACC\_G2inv ($0.9182$, $0.0167$ better than the 2nd), ACC\_G2ninv ($0.8492$, $0.0810$ better than the 2nd), weighted-F1 ($0.9483$, $0.0122$ better than the 2nd), macro-F1 ($0.8936$, $0.0412$ better than the 2nd) and AUC ($0.9776$, $0.0081$ better than the 2nd). Benefit from the CFD strategy and DMF module, our proposed framework achieves relatively high and balanced accuracy on each class. 

In addition, the statistical test results on both  datasets, as shown in Table~\ref{comparison_mrnet} and Table~\ref{comparison_men}, further indicate that our proposed framework significantly outperforms the comparison methods in most metrics.

\subsection{Ablation Analysis}
We also verify the effectiveness of the proposed CFD strategy and DMF module. Ablation studies are conducted on both adopted datasets, and the results are summarized in Table~\ref{as_mrnet} and Table~\ref{as_men}, respectively. In the tables, the models are termed as baseline1, baseline2, ..., baseline6 from the top row to the bottom row, with baseline6 representing the proposed framework. The ablation studies consider three factors: $dis\_ps$, $MoE$ and $LinG$. The $dis\_ps$ factor, which related to CFD strategy, determines whether to decouple modality-partial-shared features during the feature decoupling process. There are two factors related to the DMF module, namely $MoE$ and $LinG$. The $MoE$ factor represents whether to adopt MoE for the feature fusion, with ``$\times$" indicating fusion with concatenation operation. The $LinG$ factor denotes whether to utilize proposed LinG\_GN to generate weights in MoE, with ``$\times$" representing the use of concatenation of decoupled features as the input of the gating network. The concatenation operation for inputs in gating network is the common setting in MoE-based MML methods~\cite{xue2023dynamic,cao2023multi}.


Specifically, the order of the three modalities is PD-axial, T1-coronal and T2-sagittal on the MRNet dataset, and T1C, Flair-C, ADC on the MEN dataset.

\begin{table}[t]
\centering
\caption{The ablation study
on the MRNet dataset.}
\label{as_mrnet}
\Large
\resizebox{0.5\textwidth}{!}{
\begin{tabular}{cccccccccc}
\hline
\multicolumn{3}{c}{Proposed}   & \multirow{2}{*}{SEN} & \multirow{2}{*}{SPE} & \multirow{2}{*}{ACC} & \multirow{2}{*}{G-Mean} & \multirow{2}{*}{Ba\_ACC} & \multirow{2}{*}{AUPRC} & \multirow{2}{*}{AUC} \\ \cline{1-3}
dis\_ps & MoE    & LinG       &                      &                      &                      &                           &                          &                        &                      \\ \hline
$\times$        & $\times$ & -          & 0.6859               & 0.7157               & 0.7028               & 0.6988                    & 0.7008                   & 0.5817                 & 0.7544               \\
$\times$        & $\checkmark$     & $\times$     & 0.6859               & 0.7108               & 0.7000               & 0.6874                    & 0.6983                   & 0.5810                 & 0.7566               \\
$\times$        & $\checkmark$     & $\checkmark$  & 0.7885               & 0.6863               & 0.7306               & 0.7326                    & 0.7374                   & 0.6114                 & 0.7672               \\ \hline
$\checkmark$        & $\times$ & -          & 0.7436               & 0.7255               & 0.7333               & 0.7298                    & 0.7346                   & 0.6142                 & 0.7745               \\
$\checkmark$         & $\checkmark$     & $\times$     & 0.7820               & 0.6569               & 0.7111               & 0.7151                    & 0.7195                   & 0.5917                 & 0.7591               \\
$\checkmark$         & $\checkmark$     & $\checkmark$  & 0.7244               & 0.7500               & 0.7389               & 0.7351                    & 0.7372                   & 0.6207                 & 0.8029               \\ \hline
\end{tabular}}
\end{table}
\begin{table}[t]
\centering
\caption{The ablation study 
on the MEN dataset.}
\label{as_men}
\Large
\resizebox{0.5\textwidth}{!}{
\begin{tabular}{cccccccccc}
\hline
\multicolumn{3}{c}{Proposed}   & \multirow{2}{*}{ACC} & \multirow{2}{*}{\makecell[c]{\vspace{-1mm} ACC\\ \_G1}} & \multirow{2}{*}{\makecell[c]{\vspace{-1mm} ACC\\ \_G2inv}} & \multirow{2}{*}{\makecell[c]{\vspace{-1mm} ACC\\ \_G2ninv}} & \multirow{2}{*}{\makecell[c]{\vspace{-1mm} weighted\\ -F1}} & \multirow{2}{*}{\makecell[c]{\vspace{-1mm} macro\\ -F1}} & \multirow{2}{*}{AUC} \\ \cline{1-3}
dis\_ps & MoE    & LinG       &                      &                          &                             &                                &                               &                            &                      \\ \hline
$\times$        & $\times$ & -          & 0.9076               & 0.9371                   & 0.9333                      & 0.6635                         & 0.9113                        & 0.8091                     & 0.9584               \\
$\times$        & $\checkmark$    & $\times$     & 0.9265               & 0.9539                   & 0.8106                      & 0.8072                         & 0.9302                        & 0.8549                     & 0.9590               \\
$\times$        & $\checkmark$    & $\checkmark$ & 0.9326               & 0.9525                   & 0.8515                      & 0.8397                         & 0.9364                        & 0.8634                     & 0.9699               \\ \hline
$\checkmark$        & $\times$ & -          & 0.9101               & 0.9232                   & 0.8833                      & 0.8278                         & 0.9171                        & 0.8355                     & 0.9638               \\
$\checkmark$        & $\checkmark$    & $\times$     & 0.9297               & 0.9491                   & 0.8364                      & 0.8270                         & 0.9314                        & 0.8563                     & 0.9745               \\
$\checkmark$        & $\checkmark$    & $\checkmark$ & 0.9462               & 0.9616                   & 0.9182                      & 0.8492                         & 0.9483                        & 0.8936                     & 0.9776               \\ \hline
\end{tabular}}
\end{table}

\begin{figure}[t]
	\centering
	\includegraphics[width=\linewidth]{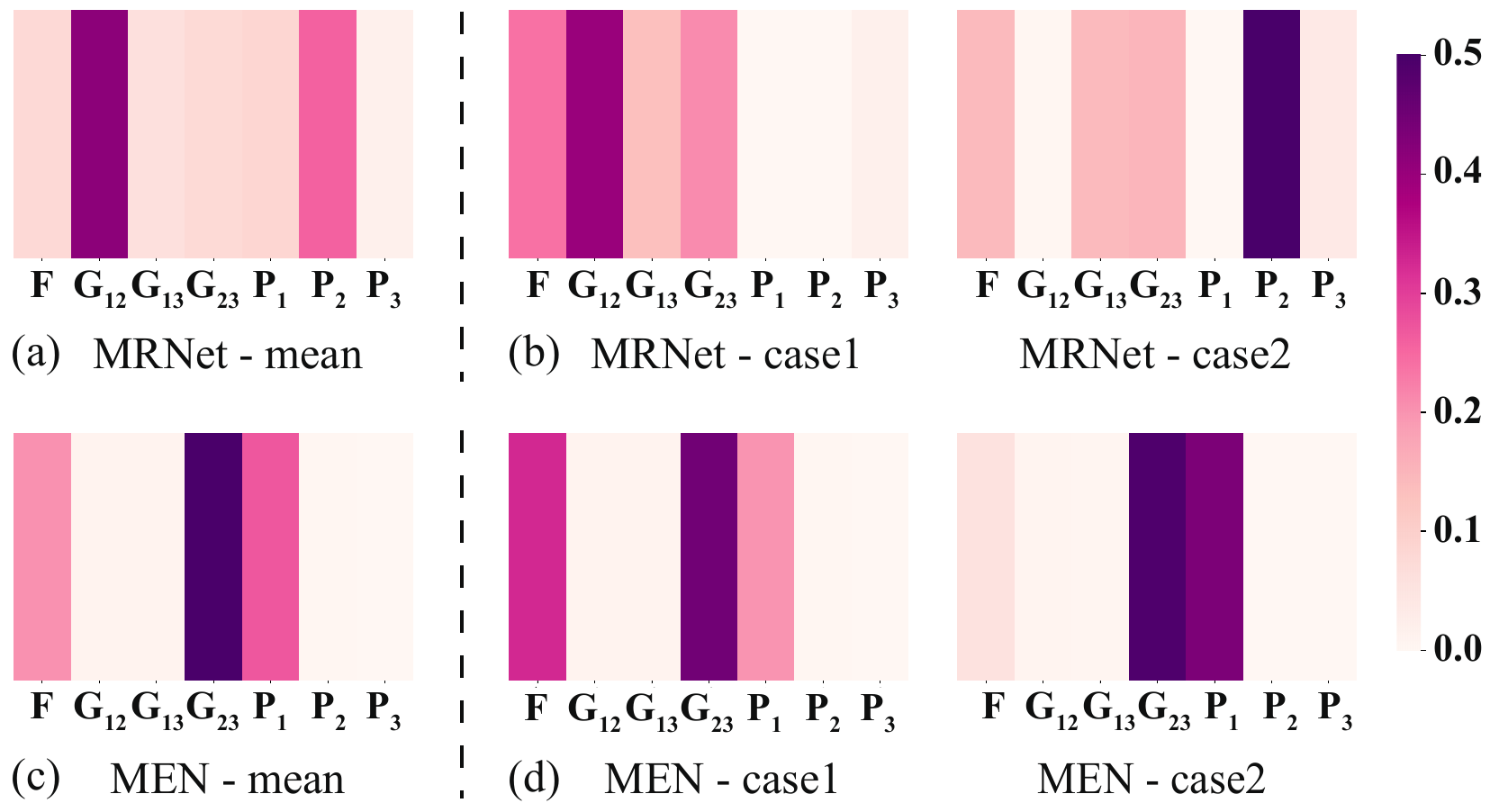}
	\caption{Heatmaps of feature weights in DMF module on the MRNet and MEN datasets. (a) and (c) show the mean feature weights of all test data on both datasets. (b) and (d) display the feature weights of two cases in each dataset, respectively.}
	\label{heatmap_moe_weights}
\end{figure}

\subsubsection{Effectiveness of the CFD Strategy}
\label{sec:CFD_visualization}
In Table~\ref{as_mrnet}, using $dis\_ps$ consistently results in significant improvements when using the same settings of factors $MoE$ and $LinG$ on the MRNet dataset (see baseline1 and baseline4, or baseline2 and baseline5, or baseline3 and baseline6).
Similar results are obtained on the MEN dataset (see Table~\ref{as_men}). These ablation studies on the $dis\_ps$ factor validate the effectiveness of the proposed CFD strategy.

We also visualize the distribution of the decoupled features using t-SNE~\cite{van2008visualizing}, and draw the heatmap which displays the cosine similarity between each pair of these features. Fig.~\ref{visual} (b) shows the visualization results for MRNet dataset. These visualization results meet the three principles which we described in Sec.~\ref{sec:CFD}.
\begin{itemize}
	\item[1)] Modality-shared features including $F_1$, $F_2$ and $F_3$ cluster together in the t-SNE visualization. These features have a cosine similarity value of $1$ with each other as shown in the corresponding heatmap.
	\item[2)] Modality-partial-shared features in each group exhibit a high degree of similarity. For example, when considering the two features ($G_{12}^1$, $G_{12}^2$) from the same group, t-SNE visualization reveals overlaps between these representations. Additionally, the heatmap shows a cosine similarity value of 1 between them.
	\item[3)] There are relatively far distance between the final modality-shared feature, each final modality-partial-shared feature and each modality-specific feature, as shown in the t-SNE visualization. These features have a vary small cosine similarity value with each other as shown in the heatmap.
\end{itemize}
Similar visualization results are observed on the MEN dataset, as depicted in Fig.~\ref{visual} (d).

Furthermore, we conduct ablation studies for $\alpha$ and $\beta$ on both adopted datasets, as shown in Fig.~\ref{visual}. Specifically, $\alpha$ and $\beta$ are balance factors of losses related to the CFD strategy ($\mathcal{L}_{sim}$ and $\mathcal{L}_{diff}$). For both datasets, the comparison results (see sub-figure (a) and (b) for MRNet dataset and sub-figure (c) and (d) for MEN dataset) clearly demonstrate that both modality-shared features and modal-partial-features are better learnt when using CFD-related losses ($\alpha\neq0$, $\beta\neq0$).


\subsubsection{Effectiveness of the DMF Module}
\label{sec:DMF_visualization}
The ablation results on the MRNet dataset are shown in Table~\ref{as_mrnet}. 
From this table, we observe that using the factor $MoE$ and $LinG$ can improve the performance when not using $dis\_ps$ (see baseline1, baseline2 and baseline3). But when using $dis\_ps$, the performance of using $MoE$ is lower than that of without $MoE$ (see baseline4 and baseline5). The possible reason is that simple concatenation used in the gating network cannot effectively capture the relationship between the final decoupled features as the number of these features increased. In contrast, our proposed LinG\_GN can dynamically capture the complex relationship between these features, allowing for better weighting and ultimately achieving improved prediction performance (see baseline4, baseline5 and baseline6). 
The ablation studies on the MEN dataset are shown in Table~\ref{as_men}.
Our proposed DMF module achieves the best performance on both MRNet and MEN datasets.


Moreover, we draw heatmaps of weights for the final decoupled features learned in the LinG\_GN. Fig.~\ref{heatmap_moe_weights} (a) and (b) show the heatmaps on the MRNet dataset. Firstly, we plot the mean weight of all cases in the test set for each final decoupled feature in a heatmap named \textit{MRNet-mean} (see Fig.~\ref{heatmap_moe_weights} (a)). This heatmap illustrates that $G_{12}$ and $P_2$ play greater roles during feature integration, with $G_{12}$ obtaining a maximum weight of around $0.4$. Specifically, $G_{12}$ represents the final modality-partial-shared feature between PD-axial and T1-corona, and $P_2$ represents modality-specific feature of T1-corona. Additionally, we randomly display the heatmaps of two cases (see Fig.~\ref{heatmap_moe_weights} (b)).

The heatmaps on the MEN dataset are shown in Fig.~\ref{heatmap_moe_weights} (c) and (d), indicating that 
$F$, $G_{23}$ and $P_1$ have more important roles during feature fusion, with $G_{23}$ obtaining the maximum weights, almost reaching 0.5. Specifically, $F$ represents the final modality-shared feature, $G_{23}$ represents the final modality-partial-shared feature between FLAIR-C and ADC, and $P_1$ represents the modality-specific feature of T1C. 

These heatmaps
demonstrate that our proposed DMF module can dynamically capture the relationships between the decoupled features across different samples (see Fig.~\ref{heatmap_moe_weights} (b) for the MRNet dataset and Fig.~\ref{heatmap_moe_weights} (d) for the MEN dataset). Moreover, there are modality-partial-shared features playing important roles during feature fusion on both datasets, further illustrating the necessity of the CFD strategy.

\section{Discussion}

\begin{figure}[t]
    \centering
    \includegraphics[width=0.9\linewidth]{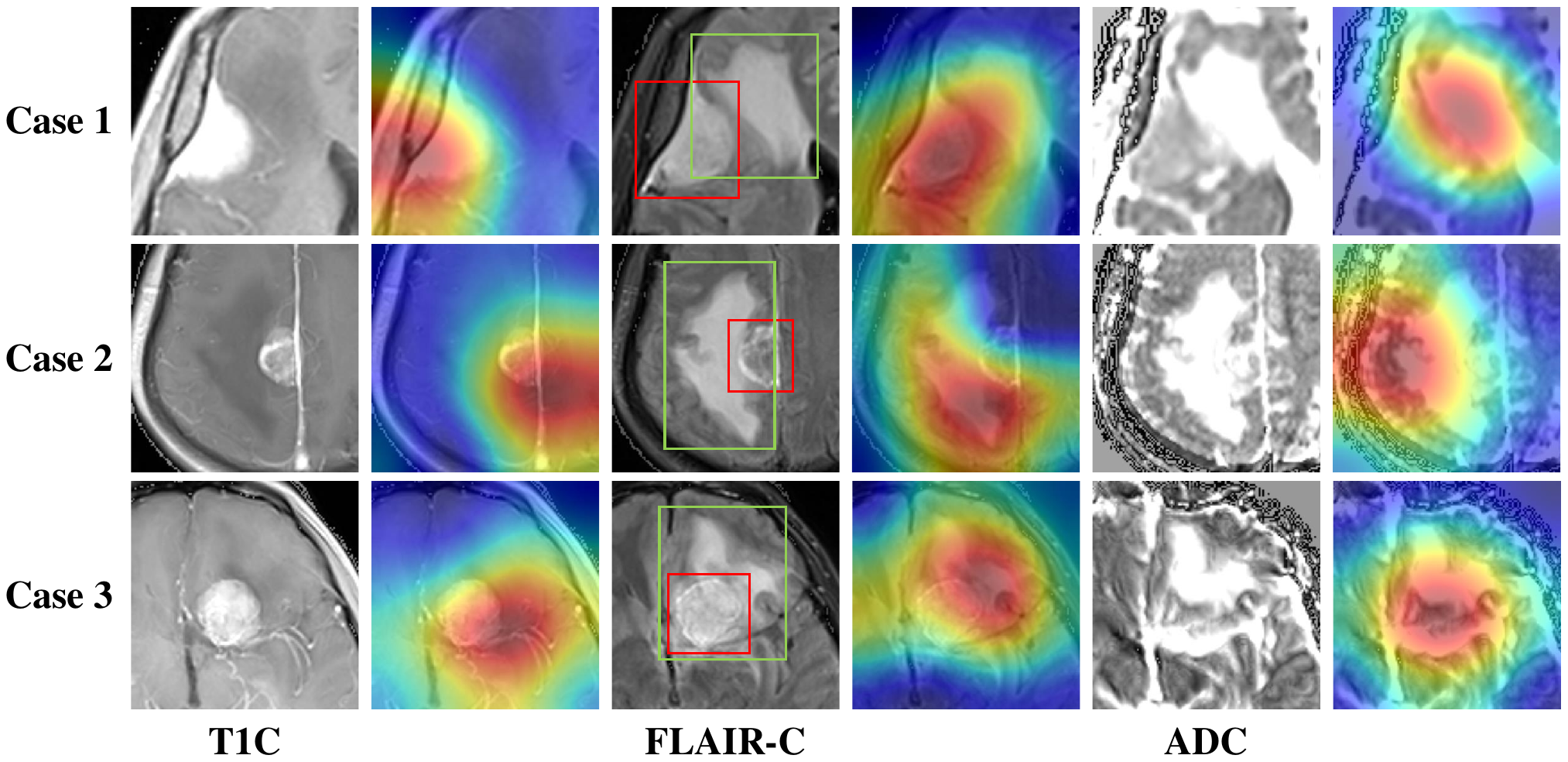}
    \caption{
    Visualization of activation maps for MRI modalities on the MEN dataset using Grad-CAM. Original images are shown on the left, with corresponding Grad-CAM images on the right. Darker red regions in the Grad-CAM images indicate areas of highest contribution to the prediction. In the FLAIR-C images, tumor and edema regions are marked with red and green boxes, respectively.}
    \label{men_cam}
\end{figure}
\subsection{Visualization analysis}
To enhance the interpretability of our framework, Grad-CAM~\cite{selvaraju2017grad} was employed to visualize the activation maps of the network for each modality in the MEN dataset. As shown in Fig~\ref{men_cam}, the model pays more attention to the tumor-edema junction across different modalities, with variations in the specific areas of focus. For the T1C modality, the model predominantly focuses on the tumor and its surrounding region. In the FLAIR-C modality, the model highlights the tumor and edema regions. Conversely, in the ADC modality, the model primarily attends to the edema region. 
Moreover, based on the feature weights in the DMF module (Fig.~\ref{heatmap_moe_weights} (c)), the characteristics of the tumor-edema junction ($F$, the final modality-shared feature), the edema ($G_{23}$, the final modality-partial-shared feature between FLAIR-C and ADC) and the tumor-surrounding region ($P1$, the modality-specific feature of T1C) contribute significantly to the final task.



\begin{figure}[t]
    \centering
    \includegraphics[width=\linewidth]{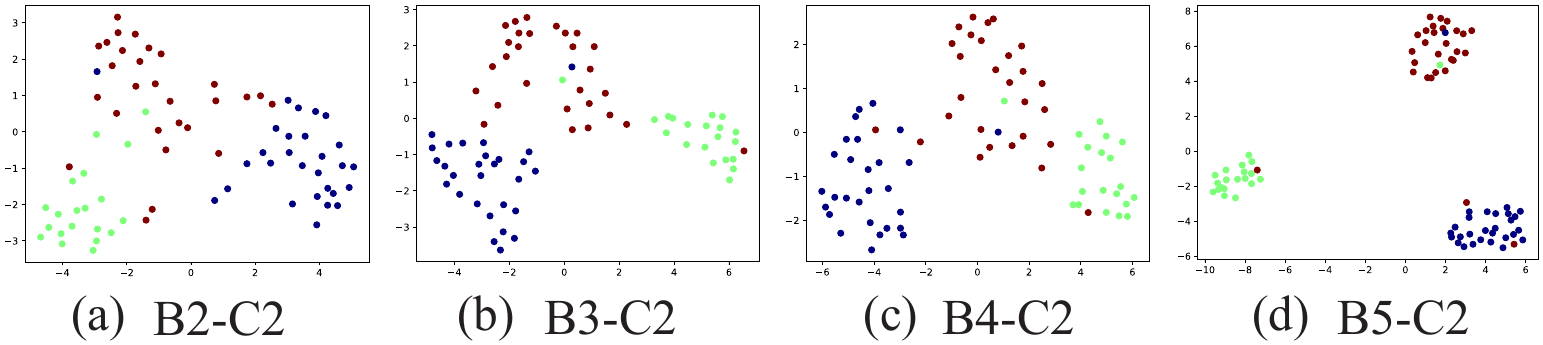}
    \caption{Data distributions in intermediate layers. The term Bm-C2 denotes the output from the $2$nd convolutional layer of the $m$th block within the ResNet18 architecture of ADC modal branch. Each point on the graph corresponds to an individual sample, with points differentiated by color to indicate various categories.}
    \label{mda_interlayer}
\end{figure}


\begin{figure}[t]
    \centering
    \includegraphics[width=\linewidth]{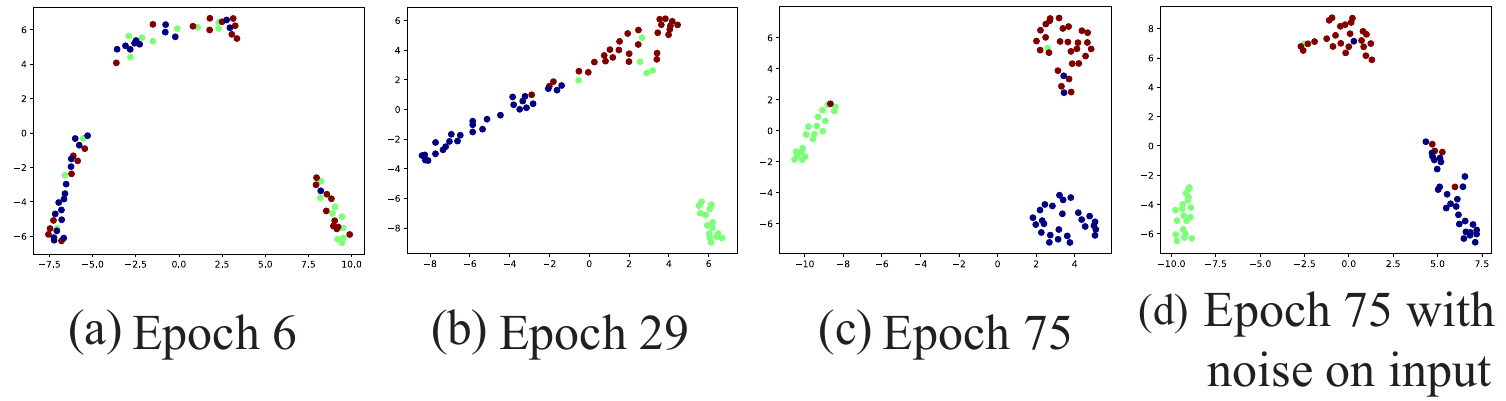}
    \caption{Data distributions in the second last FC layer with different training epochs.}
    \label{mda_multiepoch}
\end{figure}
To validate the effectiveness of our framework, we visualized the feature space distribution using the Manifold Discovery and Analysis (MDA)~\cite{islam2023revealing} algorithm on the MEN dataset, randomly selecting balanced samples across categories for optimal visualization. Fig.~\ref{mda_interlayer} demonstrates that categorical separation becomes increasingly distinct in deeper network blocks, with B2-C2 through B5-C2 showing progressive improvement in class discrimination. The temporal evolution of feature distributions across training epochs (Fig.~\ref{mda_multiepoch} (a-c)) reveals increasingly pronounced categorical boundaries. 
Furthermore, we examined the model’s robustness by analyzing feature distributions under Gaussian noise perturbation of inputs, as shown in Fig.~\ref{mda_multiepoch} (d). The results showed that the feature distributions maintained clear categorical separation, demonstrating the method’s resilience to input variations.

\subsection{Evaluation in the Four-modal Case}
\begin{figure}[t]
    \centering
    \includegraphics[width=0.85\linewidth]{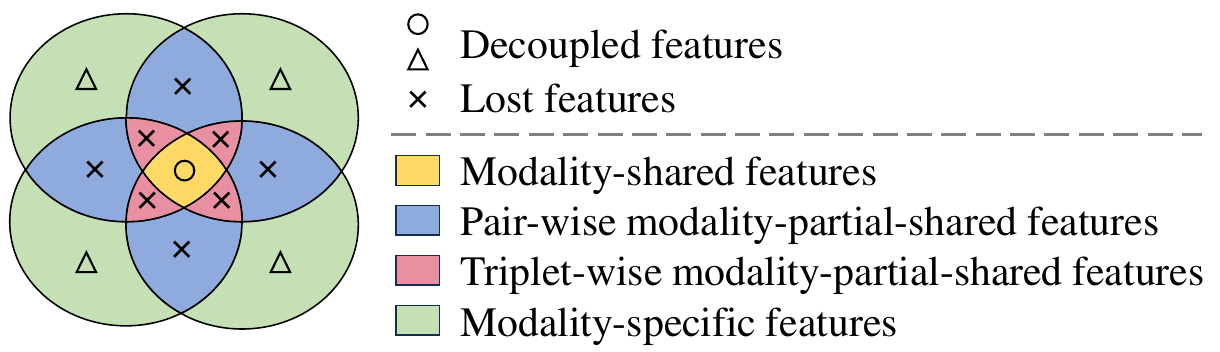}
    \caption{Illustration of incomplete feature representation on the existing FD methods in four-modal condition. Both modality-partial-shared features between pair-wise modalities (blue area) and triplet-wise modalities (red area) are lost.}
    \label{completed_information_fourmodals}
\end{figure}

In order to assess the generalization of the proposed framework on a wider range of modalities, we extend the proposed framework to MGMT promoter status prediction dataset (BraTS 2021) in the four-modal case, and the concept map is shown in Fig.~\ref{completed_information_fourmodals}. In this case, the complexity increases as we needed to decouple not only the pair-wise modality-partial-shared features but also the triplet-wise modality-partial-shared features. This leads to a total of 32 decoupled features and 15 final decoupled features. 

The comparison results on BraTS 2021 dataset are listed in Table~\ref{comparison_brats}. The seven SOTA methods obtain comparable results. Our proposed framework achieves the best results in five metrics, including ACC ($0.6137$, $0.0121$ better than the 2nd), G-Mean ($0.6089$, $0.0114$ better than the 2nd), Ba\_ACC ($0.6123$, $0.0127$ better than the 2nd), AUPRC ($0.5934$, $0.0085$ better than the 2nd), AUC ($0.6177$, $0.0173$ better than the 2nd). The statistical test results obtained on the BraTS 2021 dataset are similar to those of the MRNet and MEN datasets.

However, it is important to acknowledge that as the number of modalities increases, the relationships between them become more intricate, posing challenges for the CFD strategy. Additionally, incorporating more modalities increases the number of network parameters, further complicating network optimization. However, in clinical studies, commonly used multimodal MRI datasets typically contain two to four modalities. Our proposed framework has been designed to achieve effective performance under these conditions.

\begin{table}[t]
\centering
\caption{The comparison results on the BraTS 2021 dataset.
}
\label{comparison_brats}
\Large
\resizebox{0.5\textwidth}{!}{
\begin{tabular}{cccccccc}
\hline
Method       & SEN                                                      & SPE                                                      & ACC                                                      & G-Mean                                                  & Ba\_ACC                                                  & AUPRC                                                    & AUC                                                      \\ \hline
EmbraceNet~\cite{choi2019embracenet}   & \makecell[c]{\vspace{-1mm} 0.6034\\ \large $ \pm $ 0.1006} & \makecell[c]{\vspace{-1mm} 0.5200\\ \large $ \pm $ 0.1484} & \makecell[c]{\vspace{-1mm} 0.5639*\\ \large $ \pm $ 0.0213} & \makecell[c]{\vspace{-1mm} 0.5506*\\ \large $ \pm $ 0.0358} & \makecell[c]{\vspace{-1mm} 0.5617*\\ \large $ \pm $ 0.0269} & \makecell[c]{\vspace{-1mm} 0.5610*\\ \large $ \pm $ 0.0180} & \makecell[c]{\vspace{-1mm} 0.5942*\\ \large $ \pm $ 0.0412} \\
MISA~\cite{hazarika2020misa}         & \makecell[c]{\vspace{-1mm} 0.5515*\\ \large $ \pm $ 0.0991} & \makecell[c]{\vspace{-1mm} \textcolor{red}{0.6428}\\ \large $ \pm $ 0.0966} & \makecell[c]{\vspace{-1mm} 0.5948*\\ \large $ \pm $ 0.0288} & \makecell[c]{\vspace{-1mm} 0.5905*\\ \large $ \pm $ 0.0294} & \makecell[c]{\vspace{-1mm} 0.5972*\\ \large $ \pm $ 0.0281} & \makecell[c]{\vspace{-1mm} \textcolor{blue}{0.5849*}\\ \large $ \pm $ 0.0190} & \makecell[c]{\vspace{-1mm} 0.5992*\\ \large $ \pm $ 0.0373} \\
MAML~\cite{zhang2021modality}         & \makecell[c]{\vspace{-1mm} \textcolor{red}{0.7353}\\ \large $ \pm $ 0.1172} & \makecell[c]{\vspace{-1mm} 0.4427*\\ \large $ \pm $ 0.1313} & \makecell[c]{\vspace{-1mm} 0.5966*\\ \large $ \pm $ 0.0250} & \makecell[c]{\vspace{-1mm} 0.5607*\\ \large $ \pm $ 0.0503} & \makecell[c]{\vspace{-1mm} 0.5890*\\ \large $ \pm $ 0.0260} & \makecell[c]{\vspace{-1mm} 0.5765*\\ \large $ \pm $ 0.0164} & \makecell[c]{\vspace{-1mm} \textcolor{blue}{0.6004*}\\ \large $ \pm $ 0.0060} \\
ETMC~\cite{han2022trusted}         & \makecell[c]{\vspace{-1mm} \textcolor{blue}{0.7275}\\ \large $ \pm $ 0.1158} & \makecell[c]{\vspace{-1mm} 0.4620*\\ \large $ \pm $ 0.0974} & \makecell[c]{\vspace{-1mm} \textcolor{blue}{0.6016*}\\ \large $ \pm $ 0.0171} & \makecell[c]{\vspace{-1mm} 0.5732*\\ \large $ \pm $ 0.0154} & \makecell[c]{\vspace{-1mm} 0.5948*\\ \large $ \pm $ 0.0129} & \makecell[c]{\vspace{-1mm} 0.5798*\\ \large $ \pm $ 0.0071} & \makecell[c]{\vspace{-1mm} 0.5958*\\ \large $ \pm $ 0.0332} \\
NestedFormer~\cite{xing2022nestedformer} & \makecell[c]{\vspace{-1mm} 0.5934*\\ \large $ \pm $ 0.1240} & \makecell[c]{\vspace{-1mm} \textcolor{blue}{0.6037}\\ \large $ \pm $ 0.0879} & \makecell[c]{\vspace{-1mm} 0.5982*\\ \large $ \pm $ 0.0237} & \makecell[c]{\vspace{-1mm} 0.5921*\\ \large $ \pm $ 0.0214} & \makecell[c]{\vspace{-1mm} 0.5985*\\ \large $ \pm $ 0.0182} & \makecell[c]{\vspace{-1mm} 0.5842*\\ \large $ \pm $ 0.0104} & \makecell[c]{\vspace{-1mm} 0.5962*\\ \large $ \pm $ 0.0236} \\
ADCCA~\cite{zhou2023attentive} & \makecell[c]{\vspace{-1mm} 0.6390\\ \large $ \pm $ 0.0446} & \makecell[c]{\vspace{-1mm} 0.5602\\ \large $ \pm $ 0.0383} & \makecell[c]{\vspace{-1mm} \textcolor{blue}{0.6016*}\\ \large $ \pm $ 0.0171} & \makecell[c]{\vspace{-1mm} \textcolor{blue}{0.5975*}\\ \large $ \pm $ 0.0158} & \makecell[c]{\vspace{-1mm} \textcolor{blue}{0.5996*}\\ \large $ \pm $ 0.0167} & \makecell[c]{\vspace{-1mm} 0.5843*\\ \large $ \pm $ 0.0107} & \makecell[c]{\vspace{-1mm} 0.6003*\\ \large $ \pm $ 0.0303} \\
DMD~\cite{li2023decoupled}          & \makecell[c]{\vspace{-1mm} 0.6654\\ \large $ \pm $ 0.0229} & \makecell[c]{\vspace{-1mm} 0.5059*\\ \large $ \pm $ 0.0604} & \makecell[c]{\vspace{-1mm} 0.5898*\\ \large $ \pm $ 0.0223} & \makecell[c]{\vspace{-1mm} 0.5792*\\ \large $ \pm $ 0.0293} & \makecell[c]{\vspace{-1mm} 0.5857*\\ \large $ \pm $ 0.0239} & \makecell[c]{\vspace{-1mm} 0.5750*\\ \large $ \pm $ 0.0160} & \makecell[c]{\vspace{-1mm} 0.5686*\\ \large $ \pm $ 0.0159} \\ 
CCML~\cite{liu2024dynamic}          
& \makecell[c]{\vspace{-1mm} 0.5998\\ \large $ \pm $ 0.0335} 
& \makecell[c]{\vspace{-1mm} 0.5711\\ \large $ \pm $ 0.0225} 
& \makecell[c]{\vspace{-1mm} 0.5861*\\ \large $ \pm $ 0.0072} 
& \makecell[c]{\vspace{-1mm} 0.5848*\\ \large $ \pm $ 0.0053} 
& \makecell[c]{\vspace{-1mm} 0.5854*\\ \large $ \pm $ 0.0057} 
& \makecell[c]{\vspace{-1mm} 0.5751*\\ \large $ \pm $ 0.0032} 
& \makecell[c]{\vspace{-1mm} 0.5965*\\ \large $ \pm $ 0.0295} \\ 
GLoMo~\cite{zhuang2024glomo}          
& \makecell[c]{\vspace{-1mm} 0.6781\\ \large $ \pm $ 0.1069} 
& \makecell[c]{\vspace{-1mm} 0.4628*\\ \large $ \pm $ 0.1495} 
& \makecell[c]{\vspace{-1mm} 0.5759*\\ \large $ \pm $ 0.0155} 
& \makecell[c]{\vspace{-1mm} 0.5495*\\ \large $ \pm $ 0.0431} 
& \makecell[c]{\vspace{-1mm} 0.5704*\\ \large $ \pm $ 0.0218} 
& \makecell[c]{\vspace{-1mm} 0.5656*\\ \large $ \pm $ 0.0150} 
& \makecell[c]{\vspace{-1mm} 0.5757*\\ \large $ \pm $ 0.0124} \\ 
\hline
Proposed     & \makecell[c]{\vspace{-1mm} 0.6397\\ \large $ \pm $ 0.0605} & \makecell[c]{\vspace{-1mm} 0.5849\\ \large $ \pm $ 0.0806} & \makecell[c]{\vspace{-1mm} \textcolor{red}{0.6137}\\ \large $ \pm $ 0.0075} & \makecell[c]{\vspace{-1mm} \textcolor{red}{0.6089}\\ \large $ \pm $ 0.0136} & \makecell[c]{\vspace{-1mm} \textcolor{red}{0.6123}\\ \large $ \pm $ 0.0108} & \makecell[c]{\vspace{-1mm} \textcolor{red}{0.5934}\\ \large $ \pm $ 0.0089} & \makecell[c]{\vspace{-1mm} \textcolor{red}{0.6177}\\ \large $ \pm $ 0.0205} \\ \hline
\end{tabular}}
\end{table}
\begin{figure}[t]
	\centering
	\includegraphics[width=\linewidth]{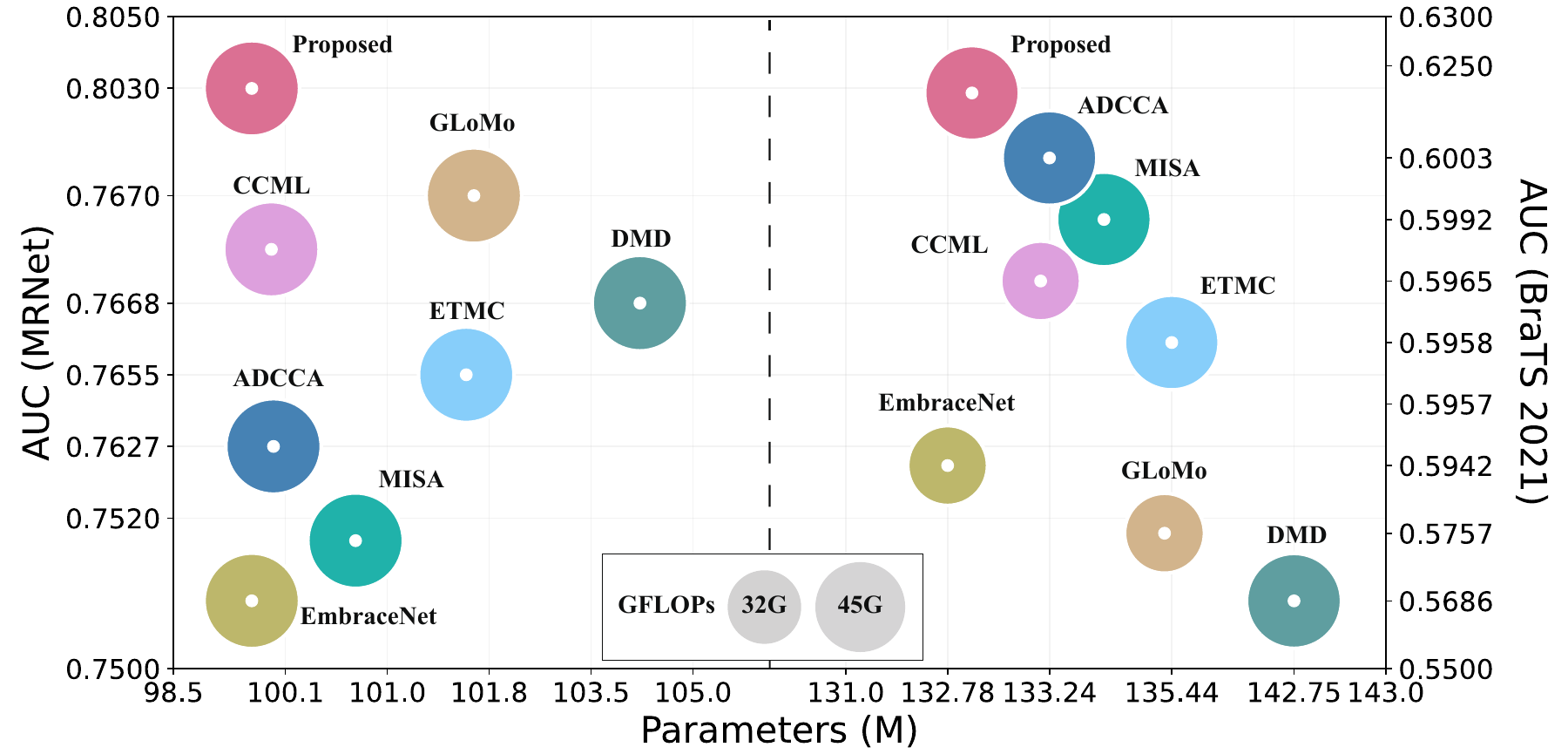}
	\caption{The computational complexity comparison including parameter and GFLOPs between proposed and comparison methods in the three-modal case (left part) and four-modal case (right part).}
	\label{vis_BC}
\end{figure}
\subsection{Computational Complexity Analysis}
As shown in Fig.~\ref{vis_BC}, we compare the number of parameters and GFLOPs between the proposed method and the comparison methods\footnote{We compare only with methods that utilize the same backbone as the proposed approach. For clarity, MAML is not shown in Fig.~\ref{vis_BC}, as its parameter and GFLOPs values are significantly larger than those of other methods. Specifically, in the three-modality case, MAML has 202.61M parameters and 165.63 GFLOPs, while in the four-modality case, these values increase to 270.09M parameters and 167.89 GFLOPs.}. In the three-modal case (left part of Fig.~\ref{vis_BC}), using MRNet as an example, the proposed method achieves the best performance with the fewest parameters and the lowest GFLOPs. In the four-modal case (right part of Fig.~\ref{vis_BC}), while EmbraceNet and CCML have the fewest parameters and GFLOPs, our method achieves a significant performance improvement with a comparable number of parameters and relatively low GFLOPs. Although the number of parameters increases for all methods from the three-modal case to the four-modal case, our proposed method achieves superior performance with only a modest increase in parameters.
\section{Conclusion and Future Work}

In this paper, we propose an effective MML framework called CFDL, incorporating a novel CFD strategy that separates multimodal information into modality-shared, modality-specific, and modality-partial-shared features, the last of which has been overlooked in previous FD-based methods. Our analysis and experiments demonstrate the critical role of modality-partial-shared features in prediction. Additionally, we present the DMF module, which explicitly and dynamically fuses the decoupled features. The LinG\_GN within the DMF module can generate the decoupled feature weights by capturing their local-global relationships. This customized fusion module can provide interpretability for clinical analysis, enabling a deeper understanding of the characteristics and behaviors of each decoupled features. Furthermore, we consider that the underlying
principles of proposed framework can be extended to other medical imaging tasks. In the future, we plan to explore the application of our framework to medical segmentation tasks, which are closely related to medical classification tasks.

\section*{References}
\bibliographystyle{IEEEtran}
\bibliography{reference.bib}

\end{document}